\documentclass[a4paper]{elsarticle}

\usepackage{url}
\usepackage{times}
\usepackage{epsfig}
\usepackage{graphicx}
\usepackage{float}
\usepackage{amsmath}
\usepackage{amssymb}

\usepackage{tabu}
\usepackage{multirow}
\usepackage{makecell}
\usepackage{subfigure}
\usepackage{color,xcolor}
\usepackage{setspace}
\usepackage{booktabs}
\usepackage[switch]{lineno} 
\usepackage{color}
\definecolor{modify}{rgb}{0, 0, 1}

\begin{document}
\begin{sloppypar}

\title{Progressive Confident Masking Attention Network for Audio-Visual Segmentation}                      



\author[1]{Yuxuan Wang*}

\author[2,3]{Jinchao Zhu*$\dagger$}

\author[4]{Feng Dong}

\author[1]{Shuyue Zhu}

\affiliation[1]{organization={Department of Computing, Imperial College London},
                city={London},
                postcode={SW7 2AZ}, 
                country={United Kingdom}}
                
\affiliation[2]{organization={College of Software, Nankai University},
                city={Tianjin},
                postcode={300000}, 
                country={China}}

\affiliation[3]{organization={Department of Automation, BNRist, Tsinghua University},
                city={Beijing},
                postcode={100089}, 
                country={China}}
                
\affiliation[4]{organization={College of Artificial Intelligence, Nankai University},
                city={Tianjin},
                postcode={300000}, 
                country={China}}

\tnotetext[]{* Equal Contributions}
\tnotetext[]{$\dagger$ Corresponding author}
\tnotetext[]{
Email: yuxuan.wang123@imperial.ac.uk (Y. Wang), jczhu@mail.nankai.edu.cn (J. Zhu), fengdong@mail.nankai.edu.cn (F. Dong), shuyue.zhu23@imperial.ac.uk (S. Zhu)
}
       

        
\begin{abstract}
Audio and visual signals typically occur simultaneously, and humans possess an innate ability to correlate and synchronize information from these two modalities. 
Recently, a challenging problem known as Audio-Visual Segmentation (AVS) has emerged, intending to produce segmentation maps for sounding objects within a scene.  %
However, the methods proposed so far have not sufficiently integrated audio and visual information, and the computational costs have been extremely high. Additionally, the outputs of different stages have not been fully utilized.
To facilitate this research, we introduce a novel Progressive Confident Masking Attention Network (PMCANet). 
It leverages attention mechanisms to uncover the intrinsic correlations between audio signals and visual frames. 
Furthermore, we design an efficient and effective cross-attention module to enhance semantic perception by selecting query tokens. 
This selection is determined through confidence-driven units based on the network's multi-stage predictive outputs. 
Experiments demonstrate that our network outperforms other AVS methods while requiring less computational resources. The code is available at: \url{https://github.com/PrettyPlate/PCMANet}.
\end{abstract}




\date{}

\maketitle

\section{Introduction}

Humans perceive the world through multiple sensory modalities, including vision, hearing, tactile sense, taste, and smell. Recently, there has been a significant shift from single-modality~\cite{2021-TCSVT-WSSD}~\cite{2023-TCSVT-LDANet}~\cite{2024-TCSVT-CDIN} to multi-modality~\cite{2022-TCSVT-MoADNet}~\cite{2023-TCSVT-LAS}~\cite{2023-TCSVT-SGFNet}~\cite{2024-IJCV-AMSP} learning, which aims to enhance machine perception capabilities. One of the most common approaches involves integrating audio and visual information, leading to extensive research on audio-visual multi-modal learning methods. Unlike homogeneous multi-modalities in computer vision, such as RGB-Depth (RGB-D)~\cite{2022-TCSVT-MoADNet} and RGB-Thermal (RGB-T)~\cite{2023-TCSVT-LAS}~\cite{2023-TCSVT-SGFNet}, audio-vision is considered a heterogeneous multi-modality. Although these modalities do not share the same distribution, there exist intrinsic correlations between them. For instance, when hearing a guitar sound, one might expect to see a person holding a guitar in the scene rather than someone sitting in front of a piano. This is because humans can capture auditory and visual cues and simultaneously synthesize the information within their brains. 
However, integrating heterogeneous data poses significant challenges for neural network construction, making it crucial to explore new models in this area of research.

\begin{figure}[t]
	\centering
	\includegraphics[width=0.7\columnwidth]{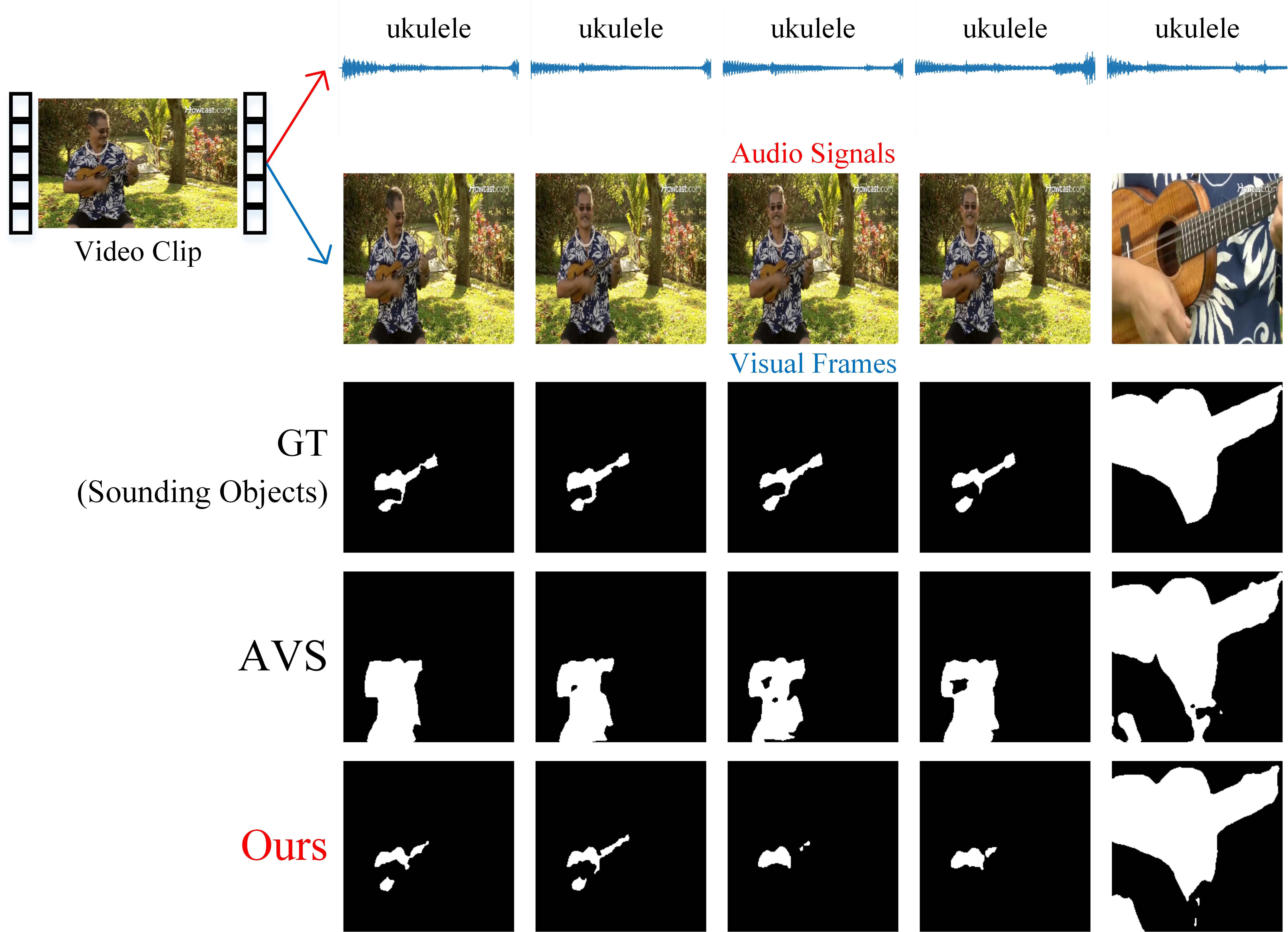}
	\caption{
		An example of the audio-visual segmentation task (S4 dataset). The video depicts a man playing the glockenspiel. Within the clip, only the glockenspiel is sounding therefore it is the only object that is labeled.
	}
	\label{illustration}
\end{figure}

In previous years, several audio-visual tasks have been proposed to enhance multi-modal perception abilities. With the advent of the big data era, millions of videos are uploaded to various applications daily, amounting to millions of hours of content in total. This has led to rapid growth in audio-visual learning, as videos serve as well-synchronized audio-visual pairs and can alleviate the need for extensive manual human labeling. Zhu \textit{et al.}~\cite{2021-IJAC-Survey} categorized discriminative audio-visual tasks into three categories: 1) separation and localization, 2) corresponding learning, and 3) representation learning. For instance, a typical audio-visual task is Audio-Visual Correspondence (AVC)~\cite{2017-ICCV-L3Net,2018-ECCV-AVENet,2018-ECCV-AV3DCNN,2018-NIPS-AVTS}, which aims to determine whether an audio signal and a video clip correspond or match each other. Another related task is Audio-Visual Event Localization (AVEL), which involves the temporal detection and localization of events within visual scenes~\cite{2018-ECCV-Unconstrained,2019-ICCV-DualAttn,2019-ICASSP-DualModal,2020-AAAI-TempIncon,2021-ACCV-InstanceAttn}. 

As mentioned earlier, the number of unlabeled videos is rapidly increasing. Research on these tasks primarily focuses on semi-supervised learning~\cite{2018-CVPR-LLSS} and unsupervised learning, particularly self-supervised learning~\cite{2016-ECCV-Ambient,2018-ECCV-SceneAnalysis,2020-ICM-LLA,2020-ECCV-Fromvideo,2020-NIPS-Discriminative,2021-NIPS-Videoparsing}. However, due to the lack of patch-level or pixel-level annotation, namely, supervision, these tasks are limited to image-level classification, and their applications are severely constrained. For years, researchers did not pay much attention to supervised learning and pixel-level annotation. Zhou \textit{et al.}~\cite{2022-ECCV-AVS, 2023-arXiv-AVSS} were the first to propose the Audio-Visual Segmentation (AVS) task and Audio-Visual Segmentation with Semantic (AVSS) task, providing benchmark datasets specifically for these tasks. 
AVS aims to segment the sounding objects in a scene by leveraging both audio and visual modalities, ensuring that the predicted segmentation mask corresponds to the source of the sound. Meanwhile, AVSS extends AVS by assigning semantic labels to segmented objects.
There is a similar task called Sound Source Localization (SSL)~\cite{2018-ECCV-SceneAnalysis,2018-CVPR-LLSS,2020-ICM-LLA,2020-ECCV-C2F,2020-ECCV-Fromvideo,2020-NIPS-Discriminative,2021-CVPR-Hardway,2022-ECCV-Easyway}, which aims to locate the regions in the visual scene that are relevant to the audio signals. Although it is not pixel-level, it is annotated at the patch-level. The results are typically represented by heat maps, which are obtained through similarity matrices or Class Activation Mapping (CAM)~\cite{2016-CVPR-CAM}.

In this paper, our main focus is on the audio-visual segmentation problem, including audio-visual semantic segmentation. Figure \ref{illustration} illustrates an example of the AVS task. 
Although significant progress has been made in previous works, three issues remain unresolved: 1. how to better integrate audio-visual features, 2. how to conserve computational resources, and 3. how to fully leverage multi-stage outputs.
We propose a novel deep neural network architecture called Progressive Confident Masking Attention Network (PCMANet). The network follows the standard encoder-decoder architecture and consists of four major components. 
We introduce an Audio-Visual Grouped Attention (AVGA) module to intuitively emphasize the visual parts that correspond to the audio signals. 
To further integrate multi-modal information, we utilize a cross-attention mechanism with a specific design to reduce computational cost while retaining the most critical information. We name this mechanism Query-Selected Cross-Attention (QSCA). 
The selection process involves generating queries using only a few tokens while masking others. This encourages the network to focus on pixels that lack confidence, resulting in reduced computational costs. 
The confidence level of each pixel is determined by the Confidence-Induced Masking (CIM) unit, which generates masks based on the phased output maps. 
These masks are propagated from deep stages to the shallower stages. 
Finally, the Guided Fusion (GF) module combines the guiding signal and features from adjacent stages to generate segmentation prediction maps. Comprehensive experiments and visualization results demonstrate the effectiveness and reasonableness of the proposed models.

Our contributions can be summarized as follows:

\begin{itemize}
	
	\item We propose an effective and efficient neural network called Progressive Confident Masking Attention Network (PCMANet) for the audio-visual segmentation task.
	
	\item We introduce an effective Audio-Visual Grouped Attention (AVGA) module to directly emphasize the visual region related to the audio signals. A novel audio-visual Query-Selected Cross-Attention (QSCA) using a considerably small number of tokens to calculate query. The selection mask is generated through a progressive Confidence-Induced Masking (CIM) unit.
	
	\item Experiments conducted on 3 audio-visual segmentation datasets illustrate that our model performs favorably against other state-of-the-art algorithms with relatively small computational costs. 
	
\end{itemize}

\section{Related Work}

\subsection{Audio-Visual Correspondence}

Audio-Visual Correspondence (AVC) is a critical task that aims to determine the relationship between audio snippets and video frames to identify whether they originate from the same video clip. 
In recent years, several approaches have been proposed to improve the performance of the AVC task. 
Arandjelovic \textit{et al.}~\cite{2017-ICCV-L3Net} proposed the L3Net which contains visual, and audio subnetworks and the fusion network. 
To further enhance the alignment of audio and visual information, the scholars developed AVENet~\cite{2018-ECCV-AVENet}, which produces aligned vision and audio embeddings as the only information source. 
Owens \textit{et al.}~\cite{2018-ECCV-AV3DCNN} utilized the 3D CNN with an early-fusion design to predict whether video frames and audio are temporally aligned. Korbar \textit{et al.}~\cite{2018-NIPS-AVTS} introduced another proxy task called Audio-Visual Temporal Synchronization (AVTS) that further considers whether a given audio sample and video clip are synchronized or not.

\subsection{Sound Source Localization}

Sound source localization (SSL) is quite similar to audio-visual segmentation. It aims to localize the regions and directions in the frames corresponding to the audio (the sound source). Senocak \textit{et al.}~\cite{2018-CVPR-LLSS} designed a two-stream network to compute the localization response by incorporating unsupervised video with a supervised loss to empower the network with prior knowledge. Cheng \textit{et al.}~\cite{2020-ICM-LLA} proposed a co-attention method employing cross-modal attention followed by self attention to learn the relationship between the audio and visual information. Chen \textit{et al.}~\cite{2021-CVPR-Hardway} introduced an automatic background mining technique and a Tri-map into the training procedure. Mo \textit{et al.}~\cite{2022-ECCV-Easyway} presented EZ-VSL, an effective multiple-instance learning framework, and proposed an object-guided localization scheme. 
Qian \textit{et al.}~\cite{2020-ECCV-C2F} utilized a two-stage learning strategy, employing a multi-task framework for classification and correspondence learning, followed by Grad-CAM~\cite{2017-ICCV-GradCAM} to generate the localization map.  
Afouras \textit{et al.}~\cite{2020-ECCV-Fromvideo} designed LWTNet to use synchronization cues to detect sound sources and group them into distinct instances. 

SSL is useful in many applications such as heading aids, television broadcasts, virtual reality, etc. 
But it doesn't have pixel-level prediction capability.

\begin{figure*}[t]
	\centering
	\includegraphics[width=1\columnwidth]{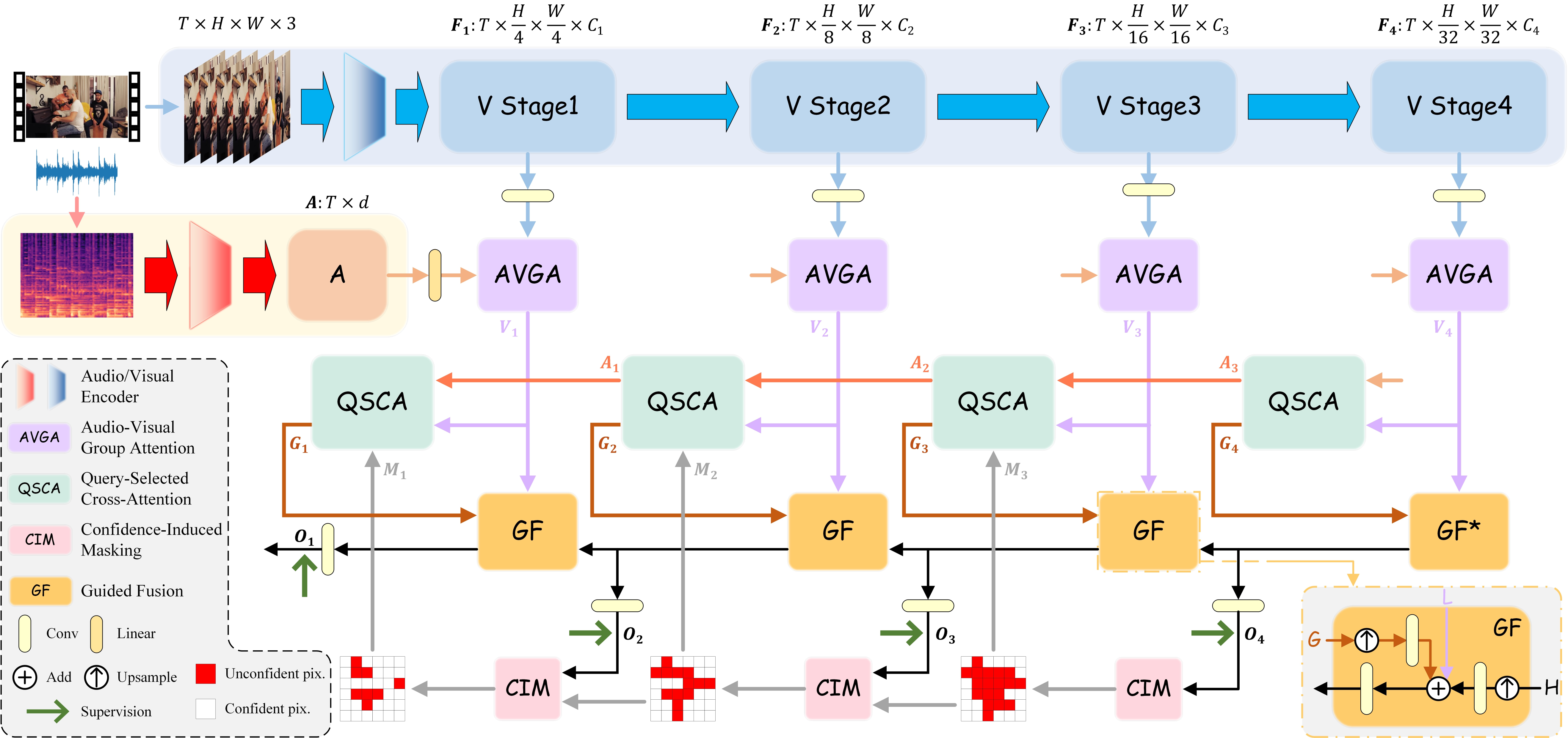}
	\caption{
		Overview of the pipeline. The network takes the video clip as input and separates it into visual frames and audio spectrograms. The outputs of the audio and visual encoders are denoted as \textbf{$F_i$} and \textbf{$A$}, respectively. The features are first integrated by AVGAs and are further processed by QSCAs to generate guide features \textbf{$G_{i}$}. Meanwhile, the multi-stage outputs, which are aggregated by GFs, will pass through CIMs to generate confidence masks and then be sent to QSCAs for further optimization. The bottom-right corner presents the internal processing logic of the GF module. The deepest GF, which does not receive the feature $H$ as input, is labeled as GF*.
	}
	\label{PMCA}   
\end{figure*}

\subsection{Audio-Visual Attention}

The attention mechanism is widely used in audio-visual tasks. Various prior works have shown its effectiveness in audio-visual learning. Xuan \textit{et al.}~\cite{2020-AAAI-TempIncon} designed a spatial, sequential, and cross-modal adaptive attention module to capture most event-related information. Wu \textit{et al.}~\cite{2019-ICCV-DualAttn} utilized a dual attention matching module to model the high-level event information while capturing local temporal information by a global cross-check mechanism. Iashin \textit{et al.}~\cite{2020-arXiv-Bimodal} proposed a Bi-modal Transformer followed by a multi-headed proposal generator to generate captions. Lin \textit{et al.}~\cite{2021-NIPS-Videoparsing} introduced a framework that learns the shared semantics through the audio and visual data across different videos and developed an audio-visual event co-occurrence module to consider the relationship of categories in audio-visual modalities. 
Lin \textit{et al.}~\cite{2021-ACCV-InstanceAttn} designed an audiovisual-Transformer to jointly encode the intra-frame and inter-frame audio-visual context. 
Nagrani \textit{et al.}~\cite{2021-NIPS-MBT} proposed a new architecture called MBT, which employs tight fusion bottlenecks to force the model to collect and condense the most relevant inputs in each modality. Sajid \textit{et al.}~\cite{2021-ICCVW-CrowdCounting} introduced an audio-visual Transformer and co-attention module to leverage the auxiliary patch-importance ranking and patch-wise crowd estimate information. Truong \textit{et al.}~\cite{2021-ICCV-Right2Talk} proposed another audio-visual Transformer to further exploit relationships across segments via a temporal self-attention mechanism. Lin \textit{et al.}~\cite{2022-arXiv-LAVISH} utilized frozen ViT backbone on both audio and visual inputs by adding trainable latent hybrid adapters to attain the cross-modal association.

\subsection{Audio-Visual Segmentation}

Audio-Visual Segmentation (AVS) task aims to generate a pixel-wise segmentation mask of the sounding objects in a scene by leveraging both audio and visual modalities. Unlike conventional segmentation tasks, AVS ensures that the segmented region corresponds to the source of the audio, distinguishing it from non-sounding objects in the scene. Audio-Visual Semantic Segmentation (AVSS) extends AVS by not only segmenting the sounding object but also assigning a semantic category to it. 

Zhou \textit{et al.}~\cite{2022-ECCV-AVS} introduced the AVS task and later extended it to AVSS~\cite{2023-arXiv-AVSS}, establishing benchmark datasets for both tasks. They designed an end-to-end framework for AVS, which adopts the TPAVI module to encode temporal pixel-wise audio-visual interaction. Hao \textit{et al.}~\cite{2024-AAAI-AVSBG} devised an efficient method utilizing bidirectional generation supervision to strengthen correlations between audio-visual modalities. Similarly, Liu \textit{et al.}~\cite{2023-MM-AVSC} introduced the audio-visual semantic correlation (AVSC) module to establish robust associations between audio and visual information. They also introduced a silent object-aware segmentation objective to address limitations in the segmentation supervision of existing AVS datasets. Compared with the one-way feature fusion approach of TPAVI, the proposed method adopts a bidirectional optimization and update strategy for audio and video feature. Compared with the traditional attention computation methods in the aforementioned approaches, the proposed approach introduces a masking and local feature update strategy. 

\section{Proposed Framework}
\subsection{Overview Structure of PCMANet}

In this section, we introduce a novel network architecture called the Progressive Confident Masking Attention Network (PCMANet), illustrated in Figure \ref{PMCA}. The visual frames undergo direct processing by the visual encoder, which can be either ResNet50~\cite{ResNet} or PVT-v2~\cite{PVT, PVTv2}. This process yields visual features denoted as $F_i$ ($i=1,2,3,4$), where $F_i\in \mathbb{R}^{T\times h_i\times w_i\times C_i}$ and $(h_i, w_i)=(H, W)/2^{(i+1)}$. Here, $T$ represents the number of frames. At the same time, the audio signal is converted into a mel spectrogram using a short-time Fourier transform, which is then passed through a convolutional neural network encoder, VGGish~\cite{VGGish}. This VGG-like model is pre-trained on AudioSet~\cite{AudioSet}. The resulting output from the audio encoder is denoted as $A\in \mathbb{R}^{T\times d}$. To ensure uniformity in channel dimensions for the audio and visual features, a simple CBR (convolution, batch normalization, and ReLU) operation is applied. 
It regularizes all visual feature channels into $C=256$.
Similarly, a linear layer is applied to the audio feature, unifying its channel dimension to $C$.

\begin{figure}[t]
	\centering
	\includegraphics[width=1\columnwidth]{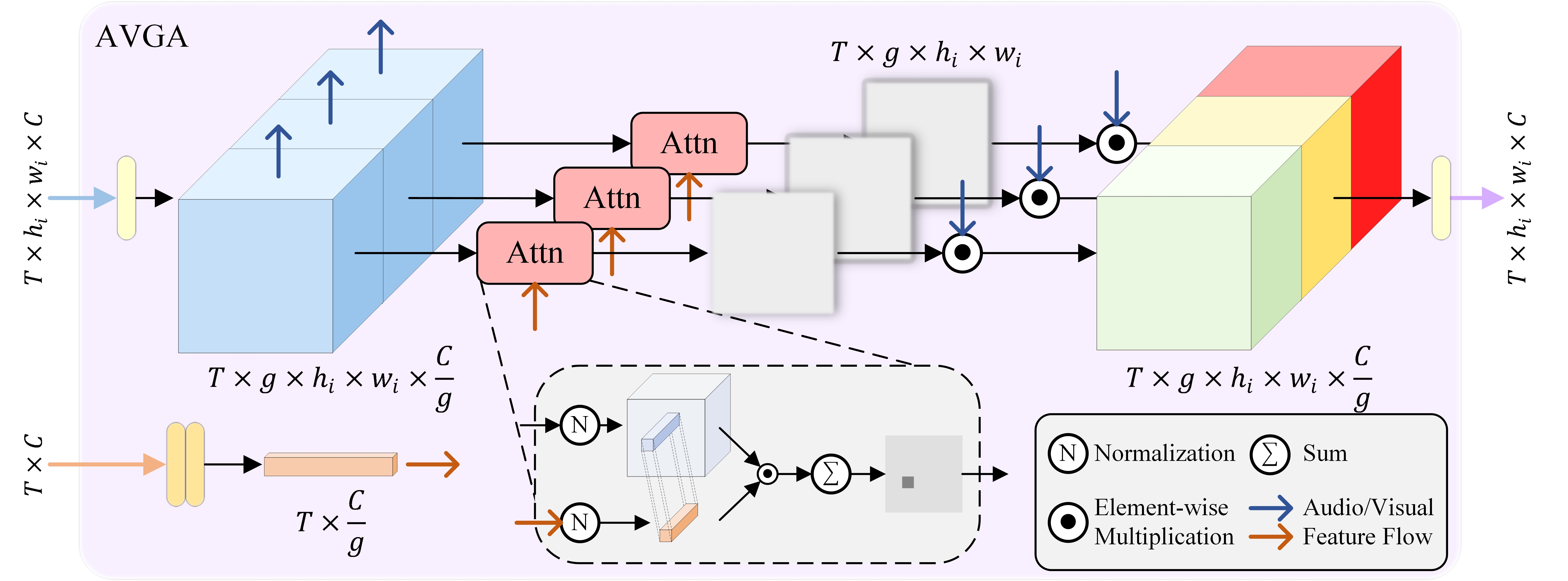}
	\caption{
		Structure of Audio-Visual Group Attention (AVGA). It integrates audio and visual features using group attention operations. The visual features are divided into multiple groups, with each group fused with audio information through the "Attn" module. This module first normalizes the input features and then performs fusion using a dot-product operation.
	}
	\label{AVGA}
\end{figure}

In the cross-modal fusion phase, the two modal features firstly are aligned at the pixel level. They are then jointly fed into the Audio-Visual Group Attention (AVGA) modules to calculate the intrinsic correlation between the two modalities with group number $g$. 
The resulting aligned features are denoted as $V_i$ ($i=1,2,3,4$). 
To further promote audio-visual information interaction, we design a Query Selection Cross Attention (QSCA) module.
This module utilizes a tailored cross-attention mechanism to better capture audio cues within visual scenes. 
It takes the audio feature and the aligned visual features, along with an additional mask, as input. 
The current QSCA module produces optimized audio features denoted as $A_i$ ($i=1,2,3$), which are used in the subsequent QSCA module. 
Besides, the QSCA module generates enhanced visual features $G_i$ ($i=1,2,3,4$), serving as guidance for the Guided Fusion (GF) decoder to produce corresponding outputs denoted as $O_i$ ($i=1,2,3,4$).

Furthermore, the cross-attention operation is exceedingly computationally intensive, and the number of attention tokens can be substantial. To address this challenge, we propose a progressive Confidence-Induced Masking (CIM) unit to identify the confident pixels that should be masked and ignored during the attention computation. The mask maps are denoted as $M_i$ ($i=1,2,3$). This is the origin of the term \textit{Query-Selected}. It directs the network's attention toward uncertain areas, typically edges, thus optimizing computational resources.

\subsection{Audio-Visual Group Attention}

With the extracted audio and visual features, it is intuitive to assess whether there exists a correlation between the audio signal and spatial localization. Previous studies\cite{2020-ECCV-Fromvideo, 2018-CVPR-LLSS} introduced an audio-visual attention mechanism by computing the cosine similarity between audio and visual features at the pixel level, and it was found to be effective in tasks such as sound source localization and separation. Drawing inspiration from these findings, we designed a novel module named Audio-Visual Group Attention (AVGA).

As depicted in Figure \ref{AVGA}, the process involves initially partitioning the visual features into groups, governed by a hyperparameter $g$, while concurrently casting the audio feature to the matching channel. Within each of these partitions, we calculate the cosine similarity by first $L_2$ normalizing both the audio and visual features, followed by computing the dot product of the audio feature with each pixel and summing the results along the channel axis. This procedure generates the attention map. Subsequently, the attention maps are employed to modulate the original features. Finally, a CBR operation is applied to consolidate the resulting output.

\subsection{Query-Selected Cross-Attention}

\begin{figure}[t]
	\centering
	\includegraphics[width=0.8\columnwidth]{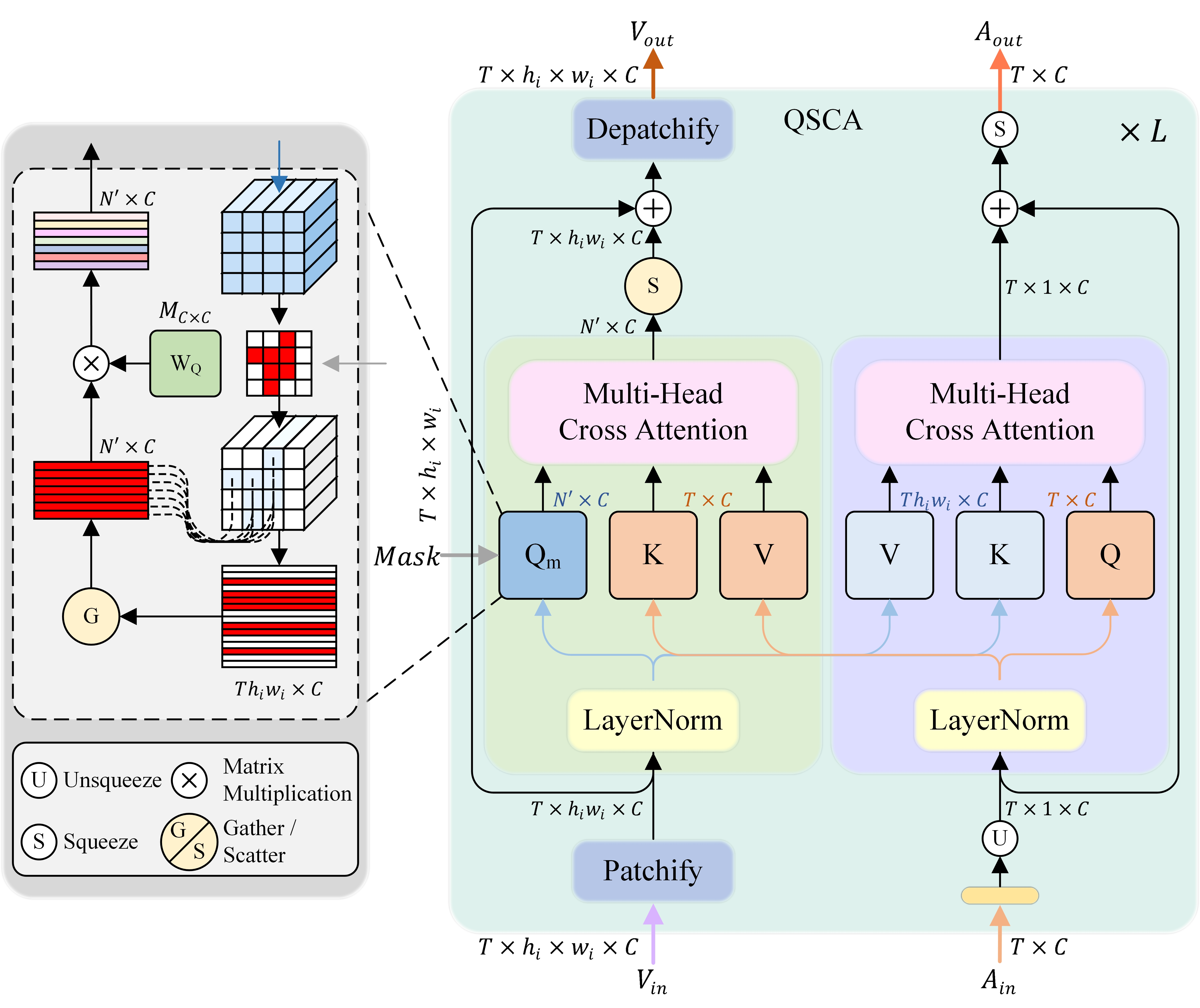}
	\caption{
		Structure of Query-Selected Cross-Attention (QSCA). 
        \textit{\textbf{Right:}} QSCA takes audio features, visual features, and an external binary mask as inputs and outputs the integrated features. Modal information is exchanged through two multi-head cross-attention blocks, with one employing query-selected attention.\quad\textit{\textbf{Left:}} Visual queries are selected using gather and scatter operations based on the mask input.
	}
	\label{QSCA}
\end{figure}

Following the initial fusion of audio-visual features, further information integration is necessary to unearth the mapping relationship between the audio signal and visual pixels. 
The Self-Attention mechanism~\cite{Transformer, ViT}, a potent tool introduced by the Transformer architecture, proves highly effective in leveraging long-range dependencies among tokens, thereby addressing the challenge of local perception limitation.
In the realm of multi-modal tasks, numerous endeavors have been undertaken based on the Transformer structure, spanning language-vision~\cite{VisualBERT, ViLBERT, VLBERT, ViLT, ALBEF, BLIP}, as well as audio-video~\cite{2021-NIPS-MBT, 2021-ACCV-InstanceAttn, 2020-arXiv-Bimodal, 2020-ICM-LLA, 2019-ICCV-DualAttn, 2021-NIPS-Videoparsing, 2021-ICCV-Right2Talk}, among others.

Inspired by prior research, we develop an innovative module named Query-Selected Cross-Attention (QSCA), as illustrated in Figure \ref{QSCA}. It takes audio and visual features as inputs and produces fused audio and visual features. In contrast to similar modules, the distinctive aspect of QSCA is its audio-guided nature. Its primary objective is to identify sounding objects within the scene, with audio serving as the guiding signal. As a result, the audio output is propagated and subsequently fed into the next QSCA module, while the visual output component is utilized in the decoder to steer the generation of the segmentation map.

Furthermore, the visual tokens are redundant, and the computational cost of the attention operation is exceedingly high. To address this challenge, we propose a new query-selected mechanism. Instead of using all tokens, the network selects tokens that lack confidence for segmentation~\cite{AdpC}. The criteria for selection are measured by the Confidence-Induced Masking unit (see Section \ref{CIM}).
\begin{figure*}[t]
	\centering
	\includegraphics[width=1\columnwidth]{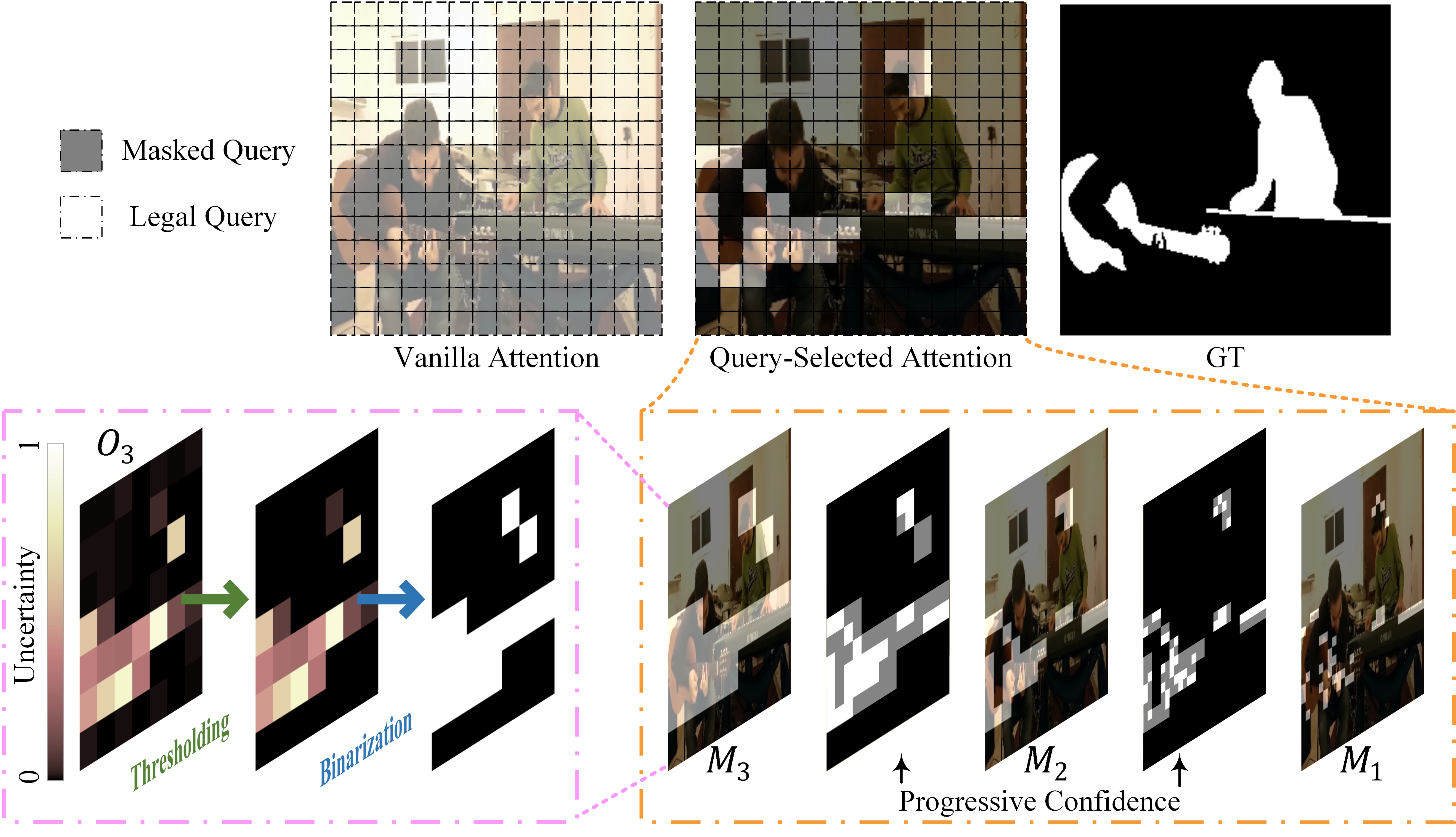}
	\caption{
		An example of the QSCA and Confidence-Induced Masking (CIM). Dark and light pixel regions represent masked and valid queries, respectively. The mask is generated from the sigmoid output, followed by thresholding and binarization. Progressive masking is achieved through iterative mask multiplication.
	}
	\label{Mask}
\end{figure*}
To regularize the representation, we denote $A_4=A$, and $M_4$ can be considered as an all-ones matrix. Thus, the module can be represented as follows.
\begin{small}
	\begin{equation}
		A_{i-1}, G_i = QSCA(A_i, V_i, M_i) \quad (i=1,2,3,4).
	\end{equation}
\end{small}

The standard attention operation can be formulated as~\cite{Transformer}:
\begin{small}
	\begin{equation}
		Attention(Q, K, V) = Softmax(\frac{QK^T}{\sqrt{d_k}}),
	\end{equation}
\end{small}
where $d_k$ is the key's dimension.

The pipeline of QSCA can be signified as: 
\begin{small}
	\begin{subequations}
		\begin{align}
			A_{i-1} &= Attention(Q_a, K_v, V_v),\\
			G_i &= Attention(Q_m, K_a, V_a),
		\end{align}
	\end{subequations}
\end{small}
where $Q_m$ is the query of visual features after selection. $Q_x, K_x, V_x$ are the query, key, and the subscript $x$ is $a$, $v$, or $m$.

$Q_m$ is calculated by the following equations:
\begin{small}
	\begin{subequations}
		\begin{align}
			V_m &= Gather(V_i * M_i),\\
			Q_m &= V_m * W_q,
		\end{align}
	\end{subequations}
\end{small}
where $V_i$ is the visual input feature and $M_i$ is the confident map generated by the CIM unit. $WQ$ is the query weight matrix. $Gather(\cdot)$ is the operation that collects all the non-zero parts of the input. Following the attention operation, the $Scatter(\cdot)$ operation is employed to put the tokens back to ensure the output's shape matches the input. 

Considering the gather-scatter operation, the computational cost is significantly reduced, and the least confident tokens are retained to help the network focus more effectively.
To elaborate, assuming there are $N$ tokens and $C$ channels, for a standard Multi-Head Self-Attention (MSA), the computational complexity can be calculated as:
\begin{small}
	\begin{equation}
		\begin{aligned}
			\Omega(\text{MSA})=4NC^2+2N^2C.
		\end{aligned}
	\end{equation}
\end{small}

While for the QSCA module, denoting the masked ratio as $r$, we can attain the complexity formula as:
\begin{small}
	\begin{equation}
		\begin{aligned}
			\Omega(\text{QSCA})=2NC^2+2N'C^2+2NN'C,
		\end{aligned}
	\end{equation}
\end{small}
where $N' = rN$, the ratio $r$ gradually decreases from $1$ to $0$. Experiments demonstrate that in the later stages of training, $r$ is lower than $10\%$, indicating that a significant portion of the tokens is masked, resulting in substantial savings in computational resources. For example, in our cases, the last QSCA module takes $28\times28$ image size input and the channel $C$ is unified to $256$. Thus, we obtain the original MSA FLOPs equals to $520.2M$. If the mask ratio exceeds $90\%$, namely $r<10\%$, the optimized FLOPs of QSCA is less than $144.3M$, which is a third of the original cross-attention computational costs. Figure \ref{Mask} provides an example of the confidence masks and the selected query.

\subsection{Confidence-Induced Masking and Guided Fusion}
\label{CIM}

\begin{figure}[t]
	\centering
	\includegraphics[width=0.6\columnwidth]{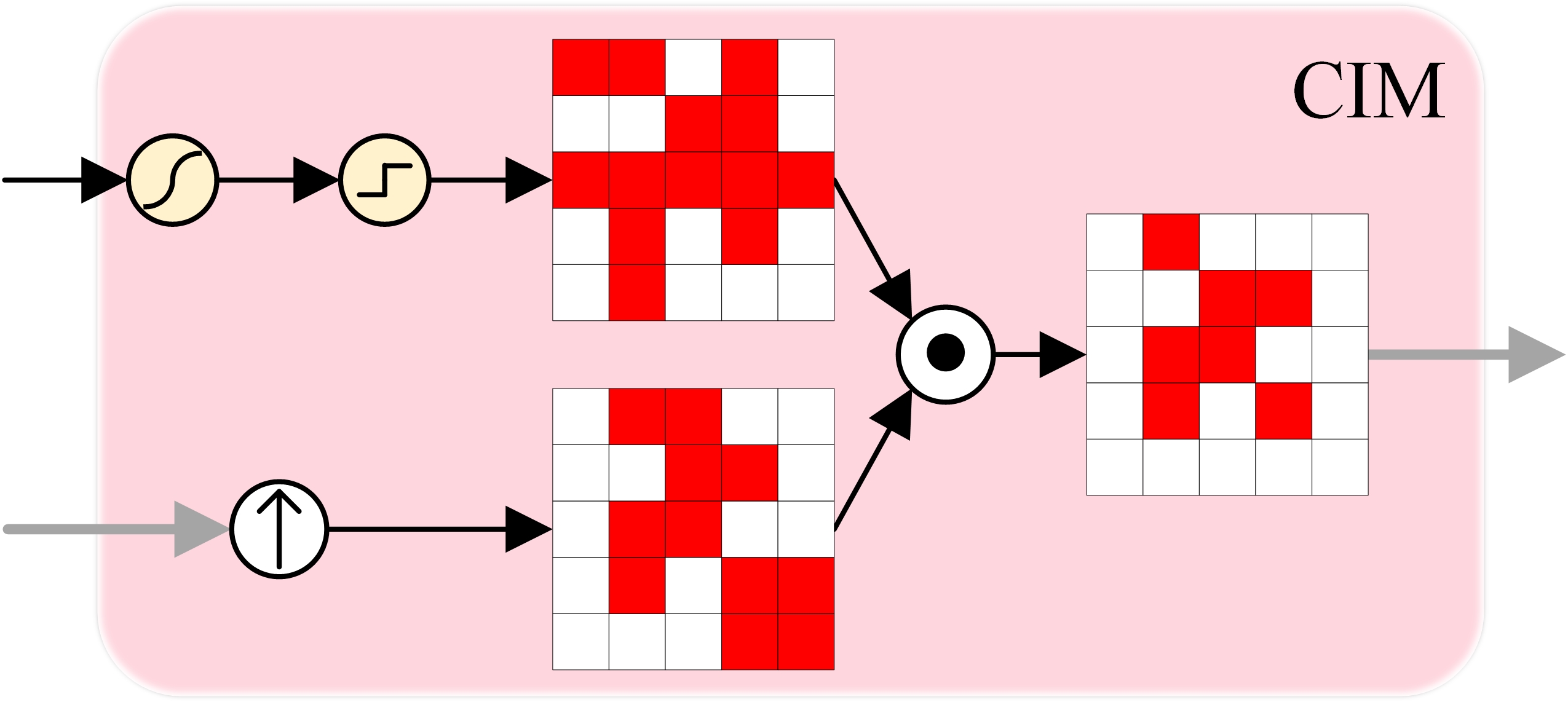}
	\caption{
		Structure of Confidence-Induced Masking (CIM). Mask is generated by the output (top arrow) of current stage and previous mask (bottom arrow).
	}
	\label{CIM_f}
\end{figure}

To determine the confidence of each token, we propose a criterion inspired by~\cite{AdpC}. Figure \ref{CIM_f} illustrates the structure of this criterion. 
Each CIM unit takes the output $O_i$ of each level Guided Fusion (GF) module as input. 
It begins by applying the $Sigmoid$ function to transform the input into a prediction map. Subsequently, it employs a $Switch$ operation with a hyperparameter $c$ to assign a value of $0$ (indicating background) to pixels falling within the range $[0, 1-c] \cup [c, 1]$, which is represented in white, and a value of $1$ (indicating lack of confidence) to pixels within the range $(1-c, c)$, depicted in red. This process selects positions that are not sufficiently confident, which are then adopted for choosing tokens in the QSCA modules.

Furthermore, we desire the network to maintain prediction consistency. This implies that once a pixel is classified as confident, it should not be subsequently categorized as unconfident in later units. To achieve this, we have devised a straightforward approach. We send the previous result to the subsequent CIM's input and combine them through pixel-wise multiplication. Consequently, the confident maps exhibit consistency and progression, encouraging the network to exercise caution in its initial predictions, which can be advantageous in a progressive structure. This process can be represented by the following equation:
\begin{small}
	\begin{equation}
		\begin{aligned}
			M_{i-1} = CIM(M_{i}, O_i).
		\end{aligned}
	\end{equation}
\end{small}

The Guided Fusion module essentially combines adjacent features ($L$, $H$), similar to FPN~\cite{FPN}. The key distinction lies in its utilization of the output feature $G$ from the QSCA module, which has already incorporated the attention information derived from audio-visual interactions. This modification ensures that the prediction generation process focuses more on the sounding objects. Notably, as shown in Figure \ref{PMCA}, the deepest GF module lacks the feature $H$ from the deeper GF module, resulting in only two inputs, $L$ and $G$, and is therefore labeled as GF*.

\section{SUPERVISION}

As we aim to regularize every confident map, we implement multi-supervision for each output $O_i$. For each supervision instance, we employ binary cross-entropy (BCE) and intersection over union (IoU) to compute the loss.

Binary cross entropy loss is widely adopted in segmentation tasks. It is defined as
\begin{small}
	\begin{equation}
		\mathcal{L}_{bce} = - \sum_{(i,j)}[G_{ij}\ln P_{ij}+(1-G_{ij})\ln (1-P_{ij})],
	\end{equation}
\end{small}where $P$ and $G$ are the prediction and the ground truth label, respectively. The subscript $ij$ refers to the pixel at $(i,j)$.

IoU loss evaluates the similarity between the prediction and Ground Truth (GT) from a holistic perspective rather than from a single pixel.
\begin{small}
	\begin{equation}
		\mathcal{L}_{iou}=1 - \frac{\sum_{(i,j)}[G_{ij}*P_{ij}]}  {\sum_{(i,j)}[P_{ij}+G_{ij}-G_{ij}*P_{ij}] }.\\
	\end{equation}
\end{small}

For single supervision, the loss function is calculated by:
\begin{small}
	\begin{equation}
		\mathcal{L}_i = \mathcal{L}^i_{bce} + \mathcal{L}^i_{iou}.
	\end{equation}
\end{small}

The final objective function is defined as:
\begin{small}
	\begin{equation}
		\mathcal{L} = \sum\limits_{i=1}^4 \lambda_i\mathcal{L}_i.
	\end{equation}
\end{small}
where $\lambda_i$ are the hyperparameters that control the proportion of each stage's loss.

\section{EXPERIMENTS}
\subsection{Datasets and Evaluation Metrics}

There are three benchmark datasets: the semi-supervised Single Sound Source Segmentation (\textbf{S4}), the fully supervised Multiple Sound Source Segmentation (\textbf{MS3}), and the fully supervised audio-visual semantic segmentation (\textbf{AVSS})~\cite{2022-ECCV-AVS,2023-arXiv-AVSS}. The S4 subset contains 4,932 videos across 23 categories, while the MS3 subset consists of 424 videos with categories selected from those in S4. These two subsets are collectively referred to as \textit{AVSBench-object}. The train/validation/test ratio for both is 70:15:15. AVSS involves a semantic segmentation task and encompasses 12,356 videos spanning 70 categories. This subset is known as \textit{AVSBench-semantic}, with a train/validation/test split of 8,498:1,304:1,554.

Videos within the \textit{AVSBench-object} subset are trimmed to 5 seconds ($T=5$), whereas those within \textit{AVSBench-semantic} are trimmed to 10 seconds ($T=10$). Videos are equally subdivided into several 1-second clips, and labels are assigned to the frames corresponding to sounding objects. Furthermore, the supervision types differ slightly. For the S4 task, videos in the training split are only annotated for the first frame, rendering it a \textit{semi-supervised} dataset. Conversely, for the MS3 and AVSS tasks, since sounding objects may change over time, videos in these datasets are fully annotated, making them \textit{fully-supervised}.

We employ two metrics to assess the performance of our model and state-of-the-art methods. The mean intersection over union (\textbf{mIoU}) evaluates the similarity of the overlapped areas. The F-measure ($\textbf{F}_\textbf{m}$)~\cite{F-measure} assesses results based on recall and precision, which can be represented as:

\begin{small}
	\begin{align}
		F_m=\frac{(1+\beta^2)\cdot precision\cdot recall}{\beta^2\cdot precision+recall},
	\end{align}
\end{small}where $\beta^2$ is set to 0.3.

\subsection{Implementation Details}

In the experimental setup, we followed the methods outlined in~\cite{2022-ECCV-AVS, 2023-arXiv-AVSS}. We employ two backbone architectures: ResNet-50~\cite{ResNet} and Pyramid Vision Transformer (PVT-v2) \cite{PVT,PVTv2} for visual feature extraction. The channel sizes for the four stages are defined as $C_{1:4} = [256, 512, 1024, 2048]$ for ResNet-50 and $C_{1:4} = [64, 128, 320, 512]$ for PVT-v2. Audio signals are initially transformed into mel spectrograms using a short-time Fourier transform. For audio feature extraction, we utilize the VGGish\cite{VGGish} model, which has been pre-trained on the AudioSet~\cite{AudioSet} dataset. Visual frames are resized to dimensions of $224\times 224$. The unified channel size is set to $C=256$, and the confidence threshold is configured at $c=0.99$. The group number of AVGA is set to $g=8$. Regarding the objective function, all $\lambda_i$ values are uniformly set to $1$.

During the training phase, we adopt the Adam optimizer with a learning rate of 1e-4. The batch size is set to 4. The number of training epochs varies, with 15, 30, and 60 epochs used for the semi-supervised S4, MS3, and the AVSS tasks, respectively. These experiments are conducted on PCs equipped with 24 GB RAM and NVIDIA RTX 3090 GPUs.

\subsection{Comparison with State of the Art Methods}
\subsubsection{Quantitative Comparison}

\begin{table*}[t]
  \centering
  \caption{Quantitative comparison on AVS dataset. The best result is highlighted in \textbf{BOLD}. The arrow $\uparrow$ of the metrics indicates the value is higher the result is better. ResNet-50, PVT-v2, and Swin refer to the ResNet-50~\cite{ResNet}, PVT-v2~\cite{PVTv2}, and Swin Transformer~\cite{Swin}, respectively.}
  \setlength{\tabcolsep}{3mm}{
    \begin{tabular}{l|c|cc|cc}
    \toprule 
    \multicolumn{1}{c|}{\multirow{2}*{Method}} & \multicolumn{1}{c|}{\multirow{2}*{Backbone}} & \multicolumn{2}{c|}{S4} & \multicolumn{2}{c}{MS3} \\
    \cmidrule(lr){3-4} \cmidrule(lr){5-6}
    &       & $\text{F}_\text{m}$$\uparrow$    & $\text{mIoU}$$\uparrow$  & $\text{F}_\text{m}$$\uparrow$    & $\text{mIoU}$$\uparrow$ \\
    \midrule
    LVS   & ResNet-50 & .510  & 37.94 & .330  & 29.45 \\
    MSSL  & ResNet-18 & .663  & 44.89 & .363  & 26.13 \\
    3DC   & ResNet-34 & .759  & 57.10 & .503  & 36.92 \\
    SST   & ResNet-101 & .801  & 66.29 & .572  & 42.57 \\
    \midrule
    TPAVI-R & ResNet-50 & .848  & 72.79 & .578  & 47.88 \\
    AVSBG & ResNet-50 & .854  & 74.13 & .568  & 44.95 \\
    PCMA-R (Ours) & ResNet-50 & \textbf{.858} & \textbf{75.24} & \textbf{.648} & \textbf{54.76} \\
    \midrule
    \midrule
    iGAN  & Swin-T & .778  & 61.59 & .544  & 42.89 \\
    LGVT  & Swin-T & .873  & 74.94 & .593  & 40.71 \\
    \midrule 
    TPAVI-T & PVT-v2 & .879  & 78.74 & .645  & 54.00 \\
    ECMMS & Swin-T & .886  & \textbf{81.29} & .657  & 59.50 \\
    PCMA-T (Ours) & PVT-v2 & \textbf{.893} & 80.10 & \textbf{.708} & \textbf{60.35} \\
    \bottomrule
    \end{tabular}%
    }
  \label{SOTA}%
\end{table*}%

\begin{table}[t]
  \centering
  \caption{Quantitative comparison on AVSS dataset. The best result is highlighted in \textbf{BOLD}.}
  \setlength{\tabcolsep}{4mm}{
    \begin{tabular}{l|c|cc}
    \toprule
    \multicolumn{1}{c|}{\multirow{2}*{Method}} & \multicolumn{1}{c|}{\multirow{2}*{Backbone}} & \multicolumn{2}{c}{AVSS} \\
    \cmidrule(lr){3-4}
    &       & $\text{F}_\text{m}$$\uparrow$   & $\text{mIoU}$$\uparrow$ \\
    \midrule
    3DC   & ResNet-34 & .216  & 17.27 \\
    \midrule
    TPAVI-R & ResNet-50 & .252  & 20.18 \\
    PCMA-R  & ResNet-50 & \textbf{.298}  & \textbf{24.88} \\
    \midrule
    \midrule
    AOT   & Swin-T   & .310  & 25.40 \\
    \midrule
    TPAVI-T & PVT-v2 & .352  & 29.77 \\
    PCMA-T  & PVT-v2 & \textbf{.361}  & \textbf{31.03} \\
    \bottomrule
    \end{tabular}%
    }
  \label{SOTA2}%
\end{table}%

To verify the performance of our method, we conducted a comparative analysis against the state-of-the-art algorithm, \textbf{TPAVI}~\cite{2022-ECCV-AVS,2023-arXiv-AVSS}, \textbf{AVSC}~\cite{2023-MM-AVSC}, \textbf{AVSBG}~\cite{2024-AAAI-AVSBG}, and three other related tasks, including sound source localization (SSL), video object segmentation (VOS), and salient object detection (SOD). Specifically, we select several typical algorithms, which are LVS~\cite{2021-CVPR-Hardway} and MSSL~\cite{2020-ECCV-C2F} for SSL, 3DC~\cite{2020-arXiv-3DC}, SST\cite{2021-CVPR-MSSL}, and AOT~\cite{2021-NIPS-AOT} for VOS, iGAN~\cite{2021-arXiv-iGAN} and LGVT~\cite{2021-NIPS-LGVT} for SOD. The quantitative results in Table \ref{SOTA} and \ref{SOTA2} demonstrate that our proposed network outperforms others, using both ResNet and Transformer backbones.

Furthermore, for a comprehensive evaluation, we measured the model sizes of all networks, as presented in Table \ref{Size}. The results highlight that PCMANet achieves superior performance with significantly reduced computational costs and faster processing speeds.

\begin{table}[t]
	\centering
	\caption{Complexity and speed comparison of different AVS algorithms. The result of AVS methods is measured by the default setting. The FPS test is conducted on a 16GB RAM, NVIDIA RTX4060 PC. The best result is highlighted in \textbf{BOLD}. The arrows $\uparrow$ ($\downarrow$) of the metrics indicate the value is higher (lower) the result is better.}
	\setlength{\tabcolsep}{1mm}{
		\begin{tabular}{l|c|ccc}  
			\toprule
		    \multicolumn{1}{c|}{Method} 
            & \multicolumn{1}{c|}{Backbone} 
            &  \multicolumn{1}{c}{Params (M)$\downarrow$}  
            & \multicolumn{1}{c}{FLOPs (G)$\downarrow$} 
            & \multicolumn{1}{c}{FPS $\uparrow$} \\
            
			\midrule
			TPAVI-R & ResNet-50
			& 91.40         & 169.320     &  13.83          \\			
			PCMA-R  & ResNet-50 & \textbf{65.68}              & \textbf{84.157}   & \textbf{32.47}          \\
            \midrule
			\midrule
			TPAVI-T & PVT-v2 & 101.32              & 154.137  &  11.63          \\			
			PCMA-T  & PVT-v2 & \textbf{94.81}              & \textbf{107.556}   &  \textbf{18.47}         \\
			\bottomrule
		\end{tabular}
	}
	\label{Size}
\end{table}

\subsubsection{Qualitative Comparison}
\begin{figure*}[t]
	\centering
	\includegraphics[width=0.9\columnwidth]{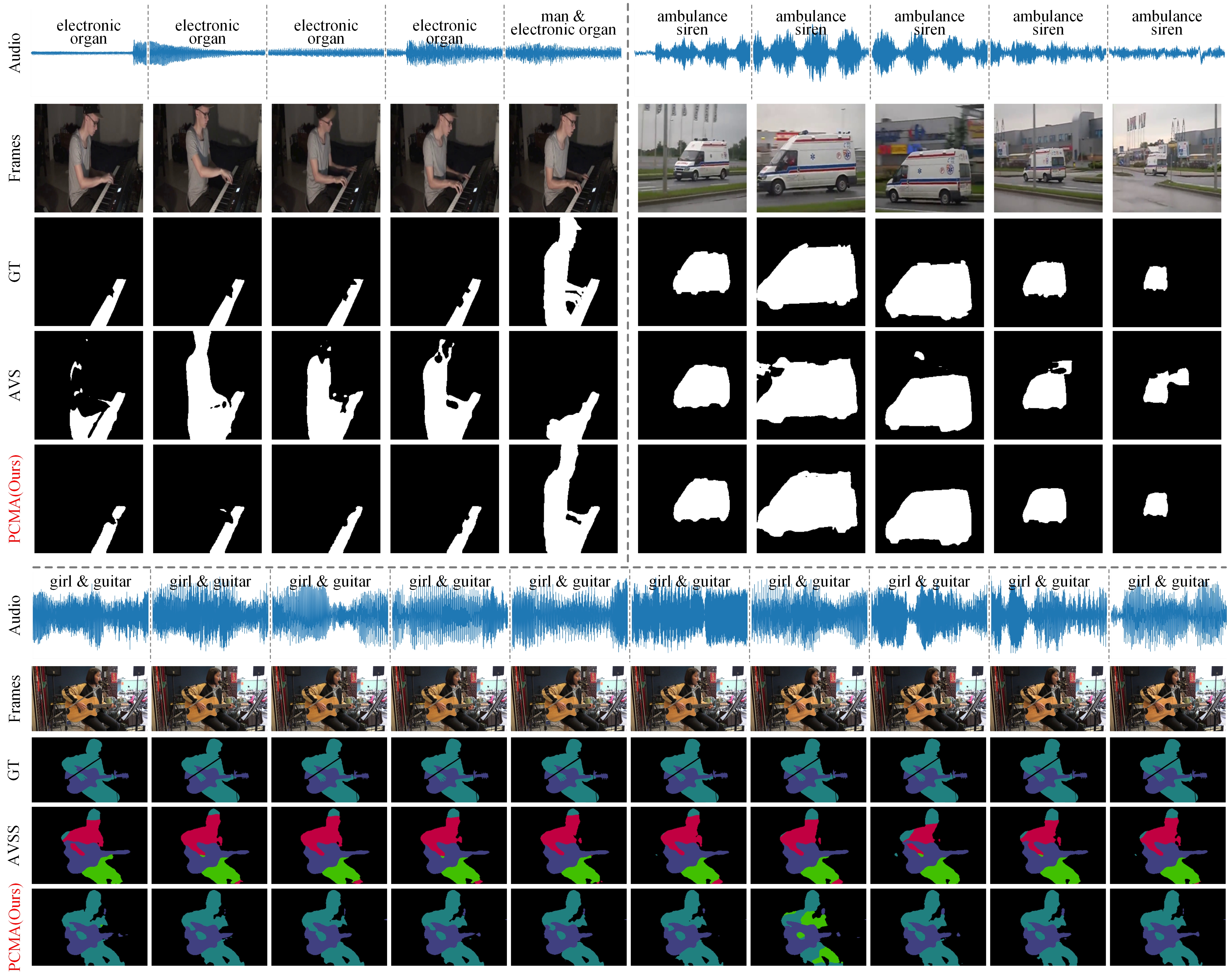}
	\caption{
		Qualitative comparison among PCMANet and other AVS and AVSS methods. (Top left: MS3 task; Top right: S4 task; Bottom: AVSS task.)
	}
	\label{Qualitative}
\end{figure*}
Figure \ref{Qualitative} presents visualizations of the qualitative results of different AVS and AVSS algorithms. It's worth noting that all predictions were generated using the ResNet-50 backbone. As depicted in the figure, in the example of the AVSS task, our method exhibits a superior ability to capture the semantics of sounding objects while preserving their integrity. In contrast, AVS mistakenly treats different parts of a girl's colorful clothing as separate objects, failing to recognize her as a complete individual.

\subsection{Ablation Analysis}
\begin{table}[t]
	\centering
	\caption{Ablation analysis of PCMANet. The best result is highlighted in \textbf{BOLD}. The abbreviations "GA", "CA", and "CI" denote the AVGA, QSCA, and CIM modules, respectively. The subscripts "M", "I", and "P" represent "Mask", "Interaction", and "Progression", respectively.
	}
	\setlength{\tabcolsep}{1mm}{
		\begin{tabular}{c|c|ccc|cc|cc}
			\toprule
            \multirow{2}*{Method} &
			\multicolumn{6}{c|}{Module} &
			\multicolumn{2}{c}{S4} \\
            
			\cmidrule(lr){2-7}\cmidrule{8-9}   
			& \multicolumn{1}{c|}{GA} & \multicolumn{1}{c}{$\text{CA}_{\text{w/o M}}$} & \multicolumn{1}{c}{$\text{CA}_{\text{w/o I}}$} & \multicolumn{1}{c|}{CA} & \multicolumn{1}{c}{$\text{CI}_{\text{w/o P}}$} & \multicolumn{1}{c|}{CI} &
			\multicolumn{1}{c}{$\text{F}_\text{m}$$\uparrow$} &  \multicolumn{1}{c}{$\text{mIoU}$$\uparrow$} \\
            
			\midrule
			$\mathcal{M}_0$ & &   &   &   &   &   &
			.832 &71.16\\
			
			$\mathcal{M}_1$ &\checkmark&   &   &   &   &   &
			.846 &72.67\\
			
			$\mathcal{M}_2$ &\checkmark&\checkmark   &   &   &   &   &
			.855 &74.29\\
			
			$\mathcal{M}_3$ &\checkmark&   &   &\checkmark   &\checkmark   &   &
			.854 &74.68\\
			
			$\mathcal{M}_4$ &\checkmark&   &\checkmark   &   &   &\checkmark   &
			.853 &74.25\\
			
			$\mathcal{M}_5$ &\checkmark&   &   &\checkmark   &   &\checkmark   &
			\textbf{.858} &\textbf{75.25}\\			
			\bottomrule
		\end{tabular}
	}
	\label{Ablation}
\end{table}
Furthermore, we conducted ablation experiments to validate the effectiveness of the modules. For the sake of simplicity, these ablation experiments were conducted using the ResNet-50 backbone in the S4 setting.

As depicted in Table \ref{Ablation}, we systematically disassembled the modules to assess the impact of each individual component. In the first row, we omitted all modules, and to facilitate a fair comparison, we directly combined the audio and visual features, serving as our baseline. The "QSCA without mask" condition indicates that we did not perform query selection and instead utilized all the tokens. The "QSCA without interaction" condition implies that the audio features generated from the QSCA modules were not passed to subsequent QSCA modules but remained constant using the initial audio feature $A$. Finally, "CIM without progression" signifies that the masks calculated by the CIM units were not progressive but treated as independent. The results underscore the effectiveness of the network's settings and modules, which are well-designed and contribute to its performance.
\begin{figure}[t]
	\centering
	\includegraphics[width=\columnwidth]{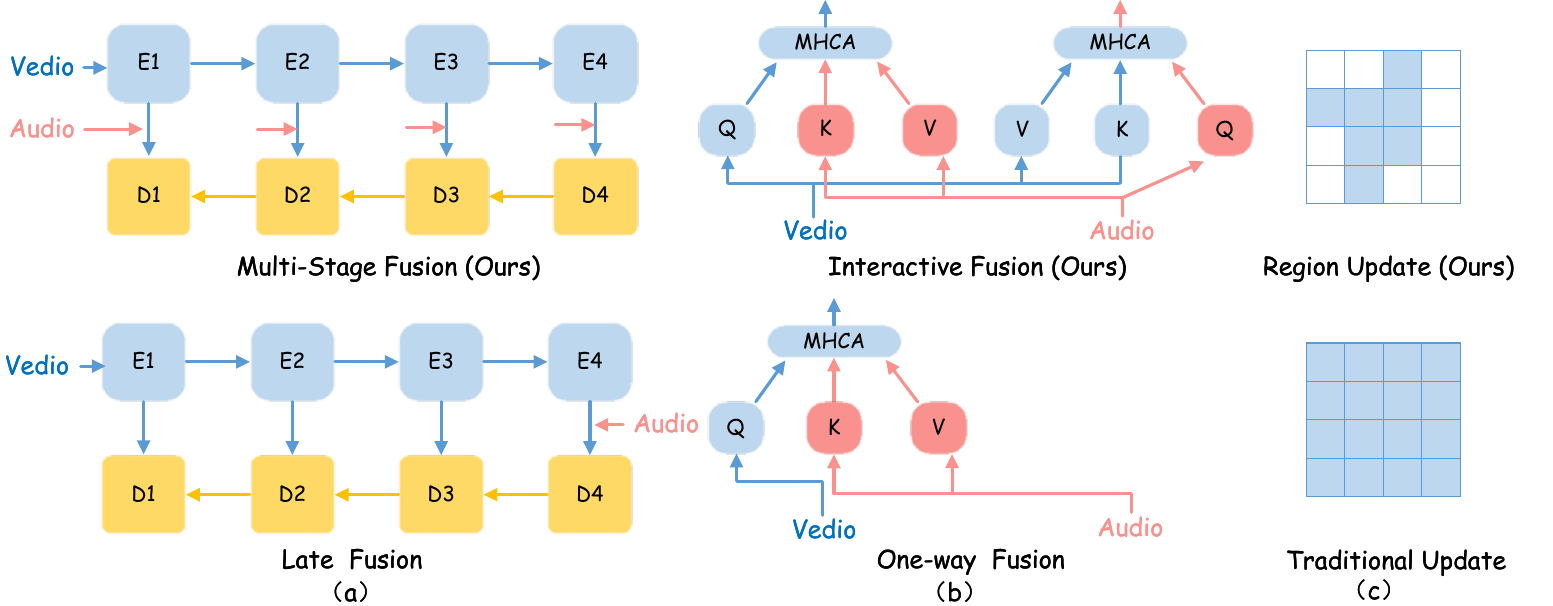}
	\caption{
        Ablation study on different fusion strategies. a) Multi-stage fusion and late fusion; b) Interactive fusion and one-way fusion; and c) Masking region update and traditional update.
	}
	\label{fusion strategies}
\end{figure}

\subsection{Study on Different Fusion Strategies}

Moreover, we explore the impact of various fusion strategies to further validate the effectiveness of our model. Specifically, we evaluate three approaches as shown in Figure \ref{fusion strategies}: 

\begin{enumerate}
    \renewcommand{\labelenumi}{\alph{enumi}.}
    \item \textbf{Comparison between multi-stage fusion (\(\mathcal{M}_1\)) and late fusion (\(\mathcal{M}_0\))}: This experiment highlights the benefit of applying AVGA fusion across all feature stages, demonstrating the advantages of a multi-stage fusion strategy over late fusion.
    \item \textbf{Comparison of interactive audio-visual fusion (\(\mathcal{M}_5\)) and single-way fusion (\(\mathcal{M}_4\))}: This analysis emphasizes the importance of bidirectional updates, showing that interactive fusion, where both audio and visual features are iteratively updated, outperforms single-way fusion. Compared to TPAVI (benchmark), which adopts a unidirectional fusion method, the bidirectional interaction approach proposed in this paper offers numerous advantages.
    \item \textbf{Comparison of the masking method (\(\mathcal{M}_5\)) and traditional methods (\(\mathcal{M}_2\))}: The results indicate that the proposed masking approach improves performance compared to conventional techniques.
\end{enumerate}

These findings collectively validate the design choices in our model and underscore the effectiveness of our proposed fusion strategies. 

\subsection{Visual Analysis}

In addition to the evaluation tests, we conducted several visualization experiments to further investigate the impact of the proposed modules. All the visualization experiments are conducted on S4 task and based on ResNet-50 backbone.

\begin{figure}[h]
	\centering
	\includegraphics[width=0.6\columnwidth]{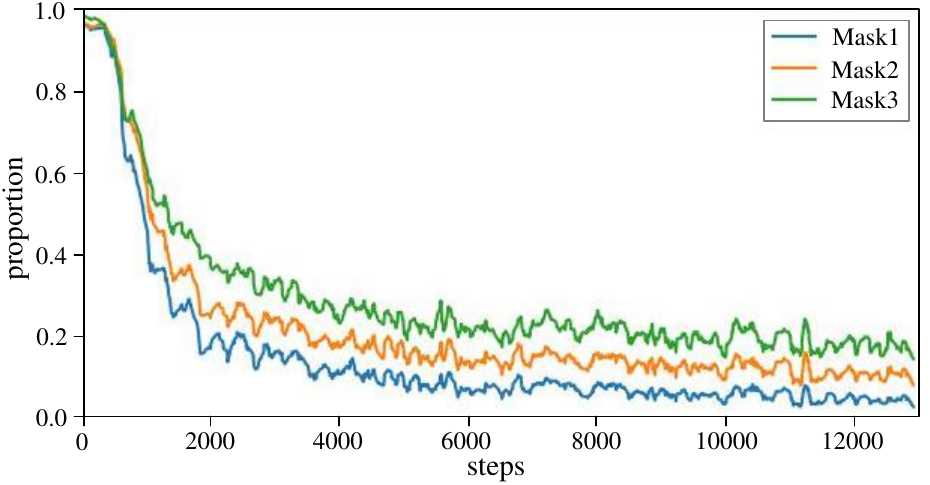}
	\caption{
		Mask trends of S4 task during the training phase.
	}
	\label{Mask_Curve}
\end{figure}

Firstly, we examined the changing trend in the proportion of masked tokens within the QSCA modules, which are derived from the CIM units. Figure \ref{Mask_Curve} illustrates the fluctuating proportions of the \textit{remaining} tokens ($1-r$) for three masks, $M_i$ (where $i=1,2,3$). It's evident that the proportions do not increase as we move from $M_1$ to $M_3$, aligning with the design to make the masks progressive. Moreover, during the initial training phase, the proportions drop rapidly and converge to approximately $10\% (M_1)$ after about one-sixth of the training period. This suggests that nearly $90\%$ of the tokens are masked after a relatively short update time, validating the efficiency and effectiveness of the modules.

Additionally, we employ visualizations of the masks to study the underlying principles of the scheme. As previously demonstrated in Figure \ref{Mask}, we can observe that the remaining pixels are primarily located along edges or in regions that are challenging to discern. The progressive setting ensures that the masked portion transitions from coarse, in the deep stage, to fine, in the shallow stage. Visualization demonstrates that the generated masks are both reasonable and efficient, ultimately leading to improved performance and reduced computational cost.

\begin{figure}[t]
	\centering
	\includegraphics[width=0.6\columnwidth]{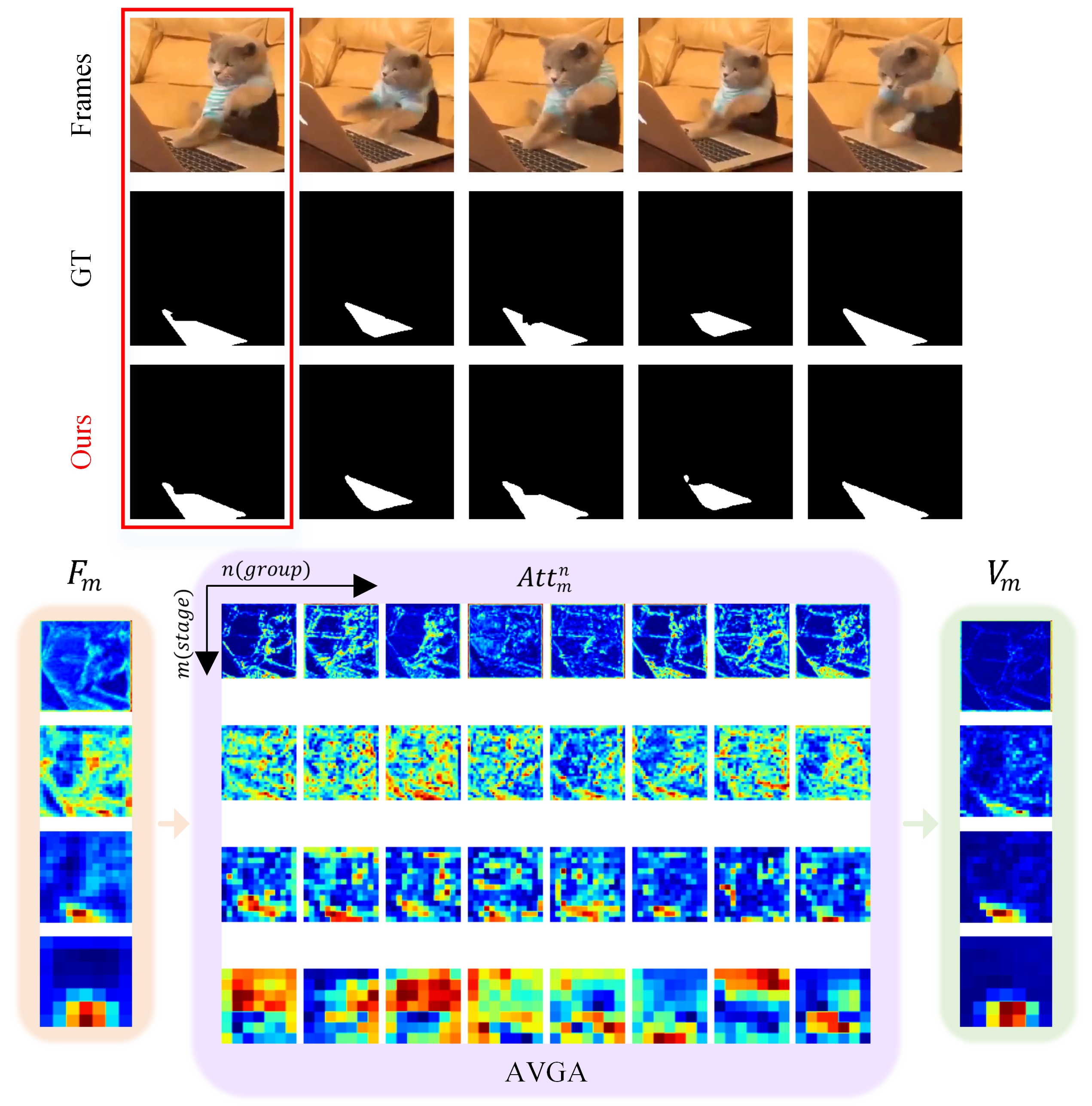}
	\caption{
		Visualization of the inputs, outputs, and attention maps of AVGA module. For the sake of simplicity, the audio wave is omitted in the figure. The $x$ and $y$ coordinate represent $n^{th}$ groups (from 1 to 8) and $m^{th}$ stages (from 1 to 4), respectively.
	}
	\label{Visualization}
\end{figure}

Finally, we delve into the AVGA modules. We denote the attention maps of AVGA as $Att_m^n$, where the subscript $m$ represents the $m^{th}$ stage ($m=1,2,3,4$), and the superscript $n$ represents the $n^{th}$ group ($n=1,2,...,g$). As illustrated in Figure \ref{Visualization}, we take $g=8$ as an example to visualize the inputs $F_m$, outputs $V_m$, and intermediate attention maps $Att_m^n$. For brevity, only the results for the first frame are displayed. In terms of the attention maps, different groups are capable of capturing feature similarities from various feature spaces. After passing through the AVGA module, it becomes evident that the features' attention is concentrated on the sounding object, such as the keyboard in this instance. This result demonstrates that the AVGA module possesses the capability to locate sounding objects and learn feature similarities from different feature perspectives.

\section{CONCLUSIONS}

This paper introduces the Progressive Confident Masking Attention Network (PCMANet) for the Audio-Visual Segmentation task. It leverages the Audio-Visual Group Attention (AVGA) mechanism to directly focus on the sounding region within the scene. Additionally, we explore a cross-attention scheme and design a novel Query-Selected Cross-Attention (QSCA) module to further integrate audio-visual features. Furthermore, to reduce the computational cost of the attention operation, we introduce Confidence-Induced Masking (CIM) units to mask a portion of the tokens based on the prediction confidence. Finally, we employ the Guided Fusion modules to generate the prediction maps of the network.

Sufficient experiments conducted on three benchmark AVS and AVSS datasets demonstrate the superior effectiveness of the proposed method compared to other algorithms. The computational efficiency and robust performance of PCMANet make it highly applicable to real-world scenarios, such as field or industrial video surveillance and edge computing systems. In the future, further investigations could focus on enhancing performance in more complex and challenging tasks, such as AVSS. 

\section*{Declaration of competing interest}
The authors declare that they have no known competing financial interests or personal relationships that could have appeared to influence the work reported in this paper.


\section*{Acknowledgments}
This work was supported by the China Postdoctoral Science Foundation under Grant 2023M741952.

\bibliographystyle{unsrt}
\bibliography{ref}

\begin{thebibliography}{10}

\bibitem{2021-TCSVT-WSSD}
Xiaoyang Zheng, Xin Tan, Jie Zhou, Lizhuang Ma, and Rynson W.~H. Lau.
\newblock Weakly-supervised saliency detection via salient object subitizing.
\newblock {\em IEEE Transactions on Circuits and Systems for Video Technology}, 31(11):4370--4380, 2021.

\bibitem{2023-TCSVT-LDANet}
Mengke Huang, Gongyang Li, Zhi Liu, and Linchao Zhu.
\newblock Lightweight distortion-aware network for salient object detection in omnidirectional images.
\newblock {\em IEEE Transactions on Circuits and Systems for Video Technology}, 33(10):6191--6197, 2023.

\bibitem{2024-TCSVT-CDIN}
Yilei Chen, Gongyang Li, Ping An, Zhi Liu, Xinpeng Huang, and Qiang Wu.
\newblock Light field salient object detection with sparse views via complementary and discriminative interaction network.
\newblock {\em IEEE Transactions on Circuits and Systems for Video Technology}, 34(2):1070--1085, 2024.

\bibitem{2022-TCSVT-MoADNet}
Xiao Jin, Kang Yi, and Jing Xu.
\newblock Moadnet: Mobile asymmetric dual-stream networks for real-time and lightweight rgb-d salient object detection.
\newblock {\em IEEE Transactions on Circuits and Systems for Video Technology}, pages 7632--7645, 2022.

\bibitem{2023-TCSVT-LAS}
Gongyang Li, Yike Wang, Zhi Liu, Xinpeng Zhang, and Dan Zeng.
\newblock Rgb-t semantic segmentation with location, activation, and sharpening.
\newblock {\em IEEE Transactions on Circuits and Systems for Video Technology}, 33(3):1223--1235, 2023.

\bibitem{2023-TCSVT-SGFNet}
Yike Wang, Gongyang Li, and Zhi Liu.
\newblock Sgfnet: Semantic-guided fusion network for rgb-thermal semantic segmentation.
\newblock {\em IEEE Transactions on Circuits and Systems for Video Technology}, pages 7737--7748, 2023.

\bibitem{2024-IJCV-AMSP}
Xiaoqi Zhao, Shijie Chang, Youwei Pang, Jiaxing Yang, Lihe Zhang, and Huchuan Lu.
\newblock Adaptive multi-source predictor for zero-shot video object segmentation.
\newblock In {\em International Journal of Computer Vision}, pages 1573--1405, 2024.

\bibitem{2021-IJAC-Survey}
Hao Zhu, Man-Di Luo, Rui Wang, Ai-Hua Zheng, and Ran He.
\newblock Deep audio-visual learning: A survey.
\newblock In {\em International Journal of Automation and Computing}, pages 351--376, 2021.

\bibitem{2017-ICCV-L3Net}
Relja Arandjelovic and Andrew Zisserman.
\newblock Look, listen and learn.
\newblock In {\em Proceedings of the IEEE International Conference on Computer Vision (ICCV)}, pages 609--617, 2017.

\bibitem{2018-ECCV-AVENet}
Relja Arandjelovic and Andrew Zissermen.
\newblock Objects that sound.
\newblock In {\em Proceedings of the European Conference on Computer Vision (ECCV)}, pages 435--451, 2018.

\bibitem{2018-ECCV-AV3DCNN}
Andrew Owens and Alexei~A Efros.
\newblock Audio-visual scene analysis with self-supervised multisensory features.
\newblock In {\em Proceedings of the European Conference on Computer Vision (ECCV)}, pages 631--648, 2018.

\bibitem{2018-NIPS-AVTS}
Bruno Korbar, Du~Tran, and Lorenzo Torresani.
\newblock Cooperative learning of audio and video models from self-supervised synchronization.
\newblock {\em Advances in Neural Information Processing Systems}, 31, 2018.

\bibitem{2018-ECCV-Unconstrained}
Yapeng Tian, Jing Shi, Bochen Li, Zhiyao Duan, and Chenliang Xu.
\newblock Audio-visual event localization in unconstrained videos.
\newblock In {\em Proceedings of the European Conference on Computer Vision (ECCV)}, pages 247--263, 2018.

\bibitem{2019-ICCV-DualAttn}
Yu~Wu, Linchao Zhu, Yan Yan, and Yi~Yang.
\newblock Dual attention matching for audio-visual event localization.
\newblock In {\em Proceedings of the IEEE International Conference on Computer Vision (ICCV)}, pages 6291--6299, 2019.

\bibitem{2019-ICASSP-DualModal}
Yan-Bo Lin, Yu-Jhe Li, and Yu-Chiang~Frank Wang.
\newblock Dual-modality seq2seq network for audio-visual event localization.
\newblock In {\em Proceedings of the International Conference on Acoustics, Speech and Signal Processing (ICASSP)}, pages 2002--2006, 2019.

\bibitem{2020-AAAI-TempIncon}
Hanyu Xuan, Zhenyu Zhang, Shuo Chen, Jian Yang, and Yan Yan.
\newblock Cross-modal attention network for temporal inconsistent audio-visual event localization.
\newblock In {\em Proceedings of the AAAI Conference on Artificial Intelligence (AAAI)}, pages 279--286, 2020.

\bibitem{2021-ACCV-InstanceAttn}
Yan-Bo Lin and Yu-Chiang~Frank Wang.
\newblock Audiovisual transformer with instance attention for audio-visual event localization.
\newblock In {\em Proceedings of the Asian Conference on Computer Vision (ACCV)}, pages 274--290. 2021.

\bibitem{2018-CVPR-LLSS}
Arda Senocak, Tae-Hyun Oh, Junsik Kim, Ming-Hsuan Yang, and In~So Kweon.
\newblock Learning to localize sound source in visual scenes.
\newblock In {\em Proceedings of the IEEE Conference on Computer Vision and Pattern Recognition (CVPR)}, pages 4358--4366, 2018.

\bibitem{2016-ECCV-Ambient}
Andrew Owens, Jiajun Wu, Josh~H McDermott, William~T Freeman, and Antonio Torralba.
\newblock Ambient sound provides supervision for visual learning.
\newblock In {\em Proceedings of the European Conference on Computer Vision (ECCV)}, pages 801--816, 2016.

\bibitem{2018-ECCV-SceneAnalysis}
Andrew Owens and Alexei~A Efros.
\newblock Audio-visual scene analysis with self-supervised multisensory features.
\newblock In {\em Proceedings of the European Conference on Computer Vision (ECCV)}, pages 631--648, 2018.

\bibitem{2020-ICM-LLA}
Ying Cheng, Ruize Wang, Zhihao Pan, Rui Feng, and Yuejie Zhang.
\newblock Look, listen, and attend: Co-attention network for self-supervised audio-visual representation learning.
\newblock In {\em Proceedings of the ACM International Conference on Multimedia}, pages 3884--3892, 2020.

\bibitem{2020-ECCV-Fromvideo}
Triantafyllos Afouras, Andrew Owens, Joon~Son Chung, and Andrew Zisserman.
\newblock Self-supervised learning of audio-visual objects from video.
\newblock In {\em Proceedings of the European Conference on Computer Vision (ECCV)}, pages 208--224, 2020.

\bibitem{2020-NIPS-Discriminative}
Di~Hu, Rui Qian, Minyue Jiang, Xiao Tan, Shilei Wen, Errui Ding, Weiyao Lin, and Dejing Dou.
\newblock Discriminative sounding objects localization via self-supervised audiovisual matching.
\newblock {\em Advances in Neural Information Processing Systems}, pages 10077--10087, 2020.

\bibitem{2021-NIPS-Videoparsing}
Yan-Bo Lin, Hung-Yu Tseng, Hsin-Ying Lee, Yen-Yu Lin, and Ming-Hsuan Yang.
\newblock Exploring cross-video and cross-modality signals for weakly-supervised audio-visual video parsing.
\newblock In {\em Proceedings of the Conference on Neural Information Processing Systems (NIPS)}, 2021.

\bibitem{2022-ECCV-AVS}
Jinxing Zhou, Jianyuan Wang, Jiayi Zhang, Weixuan Sun, Jing Zhang, Stan Birchfield, Dan Guo, Lingpeng Kong, Meng Wang, and Yiran Zhong.
\newblock Audio-visual segmentation.
\newblock In {\em Proceedings of the European Conference on Computer Vision (ECCV)}, pages 386--403, 2022.

\bibitem{2023-arXiv-AVSS}
Jinxing Zhou, Xuyang Shen, Jianyuan Wang, Jiayi Zhang, Weixuan Sun, Jing Zhang, Stan Birchfield, Dan Guo, Lingpeng Kong, Meng Wang, et~al.
\newblock Audio-visual segmentation with semantics.
\newblock {\em arXiv preprint arXiv:2301.13190}, 2023.

\bibitem{2020-ECCV-C2F}
Rui Qian, Di~Hu, Heinrich Dinkel, Mengyue Wu, Ning Xu, and Weiyao Lin.
\newblock Multiple sound sources localization from coarse to fine.
\newblock In {\em Proceedings of the European Conference on Computer Vision (ECCV)}, pages 292--308, 2020.

\bibitem{2021-CVPR-Hardway}
Honglie Chen, Weidi Xie, Triantafyllos Afouras, Arsha Nagrani, Andrea Vedaldi, and Andrew Zisserman.
\newblock Localizing visual sounds the hard way.
\newblock In {\em Proceedings of the IEEE Conference on Computer Vision and Pattern Recognition (CVPR)}, pages 16867--16876, 2021.

\bibitem{2022-ECCV-Easyway}
Shentong Mo and Pedro Morgado.
\newblock Localizing visual sounds the easy way.
\newblock In {\em Proceedings of the European Conference on Computer Vision (ECCV)}, pages 218--234, 2022.

\bibitem{2016-CVPR-CAM}
Bolei Zhou, Aditya Khosla, Agata Lapedriza, Aude Oliva, and Antonio Torralba.
\newblock Learning deep features for discriminative localization.
\newblock In {\em Proceedings of the IEEE Conference on Computer Vision and Pattern Recognition (CVPR)}, pages 2921--2929, 2016.

\bibitem{2017-ICCV-GradCAM}
Ramprasaath~R Selvaraju, Michael Cogswell, Abhishek Das, Ramakrishna Vedantam, Devi Parikh, and Dhruv Batra.
\newblock Grad-cam: Visual explanations from deep networks via gradient-based localization.
\newblock In {\em Proceedings of the IEEE International Conference on Computer Vision (ICCV)}, pages 618--626, 2017.

\bibitem{2020-arXiv-Bimodal}
Vladimir Iashin and Esa Rahtu.
\newblock A better use of audio-visual cues: Dense video captioning with bi-modal transformer.
\newblock {\em arXiv preprint arXiv:2005.08271}, 2020.

\bibitem{2021-NIPS-MBT}
Arsha Nagrani, Shan Yang, Anurag Arnab, Aren Jansen, Cordelia Schmid, and Chen Sun.
\newblock Attention bottlenecks for multimodal fusion.
\newblock In {\em Proceedings of the Annual Conference on Neural Information Processing Systems (NeurIPS)}, 2021.

\bibitem{2021-ICCVW-CrowdCounting}
Usman Sajid, Xiangyu Chen, Hasan Sajid, Taejoon Kim, and Guanghui Wang.
\newblock Audio-visual transformer based crowd counting.
\newblock In {\em Proceedings of the IEEE International Conference on Computer Vision Workshops (ICCVW)}, pages 2249--2259, 2021.

\bibitem{2021-ICCV-Right2Talk}
Thanh-Dat Truong, Chi~Nhan Duong, The De~Vu, Hoang~Anh Pham, Bhiksha Raj, Ngan Le, and Khoa Luu.
\newblock The right to talk: An audio-visual transformer approach.
\newblock In {\em Proceedings of the IEEE International Conference on Computer Vision (ICCV)}, pages 1085--1094.

\bibitem{2022-arXiv-LAVISH}
Yan-Bo Lin, Yi-Lin Sung, Jie Lei, Mohit Bansal, and Gedas Bertasius.
\newblock Vision transformers are parameter-efficient audio-visual learners.
\newblock {\em arXiv preprint arXiv:2212.07983}, 2022.

\bibitem{2024-AAAI-AVSBG}
Dawei Hao, Yuxin Mao, Bowen He, Xiaodong Han, Yuchao Dai, and Yiran Zhong.
\newblock Improving audio-visual segmentation with bidirectional generation.
\newblock In {\em Proceedings of the AAAI Conference on Artificial Intelligence}, pages 2067--2075, 2024.

\bibitem{2023-MM-AVSC}
Chen Liu, Peike~Patrick Li, Xingqun Qi, Hu~Zhang, Lincheng Li, Dadong Wang, and Xin Yu.
\newblock Audio-visual segmentation by exploring cross-modal mutual semantics.
\newblock In {\em Proceedings of the 31st ACM International Conference on Multimedia}, pages 7590--7598, 2023.

\bibitem{ResNet}
Kaiming He, Xiangyu Zhang, Shaoqing Ren, and Jian Sun.
\newblock Deep residual learning for image recognition.
\newblock In {\em Proceedings of the IEEE Conference on Computer Vision and Pattern Recognition (CVPR)}, pages 770--778, 2016.

\bibitem{PVT}
Wenhai Wang, Enze Xie, Xiang Li, Deng-Ping Fan, Kaitao Song, Ding Liang, Tong Lu, Ping Luo, and Ling Shao.
\newblock Pyramid vision transformer: A versatile backbone for dense prediction without convolutions.
\newblock In {\em Proceedings of the IEEE International Conference on Computer Vision (ICCV)}, pages 548--558, 2021.

\bibitem{PVTv2}
Wenhai Wang, Enze Xie, Xiang Li, Dengping Fan, Kaitao Song, Ding Liang, Tong Lu, Ping Luo, and Ling Shao.
\newblock Pvt v2: Improved baselines with pyramid vision transformer.
\newblock {\em Computational Visual Media}, pages 415--424, 2022.

\bibitem{VGGish}
Shawn Hershey, Sourish Chaudhuri, Daniel~PW Ellis, Jort~F Gemmeke, Aren Jansen, R~Channing Moore, Manoj Plakal, Devin Platt, Rif~A Saurous, Bryan Seybold, et~al.
\newblock Cnn architectures for large-scale audio classification.
\newblock In {\em Proceedings of the IIEEE International Conference on Acoustics, Speech and Signal Processing (ICASSP)}, pages 131--135, 2017.

\bibitem{AudioSet}
Jort~F Gemmeke, Daniel~PW Ellis, Dylan Freedman, Aren Jansen, Wade Lawrence, R~Channing Moore, Manoj Plakal, and Marvin Ritter.
\newblock Audio set: An ontology and human-labeled dataset for audio events.
\newblock In {\em Proceedings of the IEEE International Conference on Acoustics, Speech and Signal Processing (ICASSP)}, pages 776--780, 2017.

\bibitem{Transformer}
Ashish Vaswani, Noam Shazeer, Niki Parmar, Jakob Uszkoreit, Llion Jones, Aidan~N Gomez, Łukasz Kaiser, and Illia Polosukhin.
\newblock Attention is all you need.
\newblock In {\em Proceedings of the Annual Conference on Neural Information Processing Systems (NeurIPS)}, page~11.

\bibitem{ViT}
Alexey Dosovitskiy, Lucas Beyer, Alexander Kolesnikov, Dirk Weissenborn, Xiaohua Zhai, Thomas Unterthiner, Mostafa Dehghani, Matthias Minderer, Georg Heigold, Sylvain Gelly, et~al.
\newblock An image is worth 16x16 words: Transformers for image recognition at scale.
\newblock {\em arXiv preprint arXiv:2010.11929}, 2020.

\bibitem{VisualBERT}
Liunian~Harold Li, Mark Yatskar, Da~Yin, Cho-Jui Hsieh, and Kai-Wei Chang.
\newblock Visualbert: A simple and performant baseline for vision and language.
\newblock {\em arXiv preprint arXiv:1908.03557}, 2019.

\bibitem{ViLBERT}
Jiasen Lu, Dhruv Batra, Devi Parikh, and Stefan Lee.
\newblock Vilbert: Pretraining task-agnostic visiolinguistic representations for vision-and-language tasks.
\newblock In {\em Proceedings of the Annual Conference on Neural Information Processing Systems (NeurIPS)}.

\bibitem{VLBERT}
Weijie Su, Xizhou Zhu, Yue Cao, Bin Li, Lewei Lu, Furu Wei, and Jifeng Dai.
\newblock Vl-bert: Pre-training of generic visual-linguistic representations.
\newblock {\em arXiv preprint arXiv:1908.08530}, 2019.

\bibitem{ViLT}
Wonjae Kim, Bokyung Son, and Ildoo Kim.
\newblock Vilt: Vision-and-language transformer without convolution or region supervision.
\newblock In {\em Proceedings of the International Conference on Machine Learning}, pages 5583--5594.

\bibitem{ALBEF}
Junnan Li, Ramprasaath Selvaraju, Akhilesh Gotmare, Shafiq Joty, Caiming Xiong, and Steven Chu~Hong Hoi.
\newblock Align before fuse: Vision and language representation learning with momentum distillation.
\newblock In {\em Advances in Neural Information Processing Systems (NeurIPS)}, pages 9694--9705.

\bibitem{BLIP}
Junnan Li, Dongxu Li, Caiming Xiong, and Steven Hoi.
\newblock Blip: Bootstrapping language-image pre-training for unified vision-language understanding and generation.
\newblock In {\em Proceedings of the 39th International Conference on Machine Learning}, pages 12888--12900.

\bibitem{AdpC}
Zhuang Liu, Zhiqiu Xu, Hung-Ju Wang, Trevor Darrell, and Evan Shelhamer.
\newblock Anytime dense prediction with confidence adaptivity.
\newblock {\em arXiv preprint arXiv:2104.00749}, 2021.

\bibitem{FPN}
Tsung-Yi Lin, Piotr Dollar, Ross Girshick, Kaiming He, Bharath Hariharan, and Serge Belongie.
\newblock Feature pyramid networks for object detection.
\newblock In {\em Proceedings of the IEEE Conference on Computer Vision and Pattern Recognition (CVPR)}, pages 936--944.

\bibitem{F-measure}
Radhakrishna Achanta, Sheila Hemami, Francisco Estrada, and Sabine Susstrunk.
\newblock Frequency-tuned salient region detection.
\newblock In {\em Proceedings of the IEEE Conference on Computer Vision and Pattern Recognition (CVPR)}, pages 1597--1604, 2009.

\bibitem{Swin}
Ze~Liu, Yutong Lin, Yue Cao, Han Hu, Yixuan Wei, Zheng Zhang, Stephen Lin, and Baining Guo.
\newblock Swin transformer: Hierarchical vision transformer using shifted windows.
\newblock In {\em Proceedings of the IEEE international conference on computer vision}, pages 10012--10022, 2021.

\bibitem{2020-arXiv-3DC}
Sabarinath Mahadevan, Ali Athar, Aljo{\v{s}}a O{\v{s}}ep, Sebastian Hennen, Laura Leal-Taix{\'e}, and Bastian Leibe.
\newblock Making a case for 3d convolutions for object segmentation in videos.
\newblock {\em arXiv preprint arXiv:2008.11516}, 2020.

\bibitem{2021-CVPR-MSSL}
Brendan Duke, Abdalla Ahmed, Christian Wolf, Parham Aarabi, and Graham~W Taylor.
\newblock Sstvos: Sparse spatiotemporal transformers for video object segmentation.
\newblock In {\em Proceedings of the IEEE Conference on Computer Vision and Pattern Recognition (CVPR)}, pages 5912--5921, 2021.

\bibitem{2021-NIPS-AOT}
Zongxin Yang, Yunchao Wei, and Yi~Yang.
\newblock Associating objects with transformers for video object segmentation.
\newblock {\em Proceedings of the Annual Conference on Neural Information Processing Systems (NeurIPS)}, pages 2491--2502, 2021.

\bibitem{2021-arXiv-iGAN}
Yuxin Mao, Jing Zhang, Zhexiong Wan, Yuchao Dai, Aixuan Li, Yunqiu Lv, Xinyu Tian, Deng-Ping Fan, and Nick Barnes.
\newblock Transformer transforms salient object detection and camouflaged object detection.
\newblock {\em arXiv preprint arXiv:2104.10127}, 2021.

\bibitem{2021-NIPS-LGVT}
Jing Zhang, Jianwen Xie, Nick Barnes, and Ping Li.
\newblock Learning generative vision transformer with energy-based latent space for saliency prediction.
\newblock {\em Proceedings of the Annual Conference on Neural Information Processing Systems (NeurIPS)}, pages 15448--15463, 2021.

\end{thebibliography}



\end{sloppypar}

\clearpage
\appendix
\section{More Visualization}
\begin{figure}[h]
	\centering
	\includegraphics[width=\columnwidth]{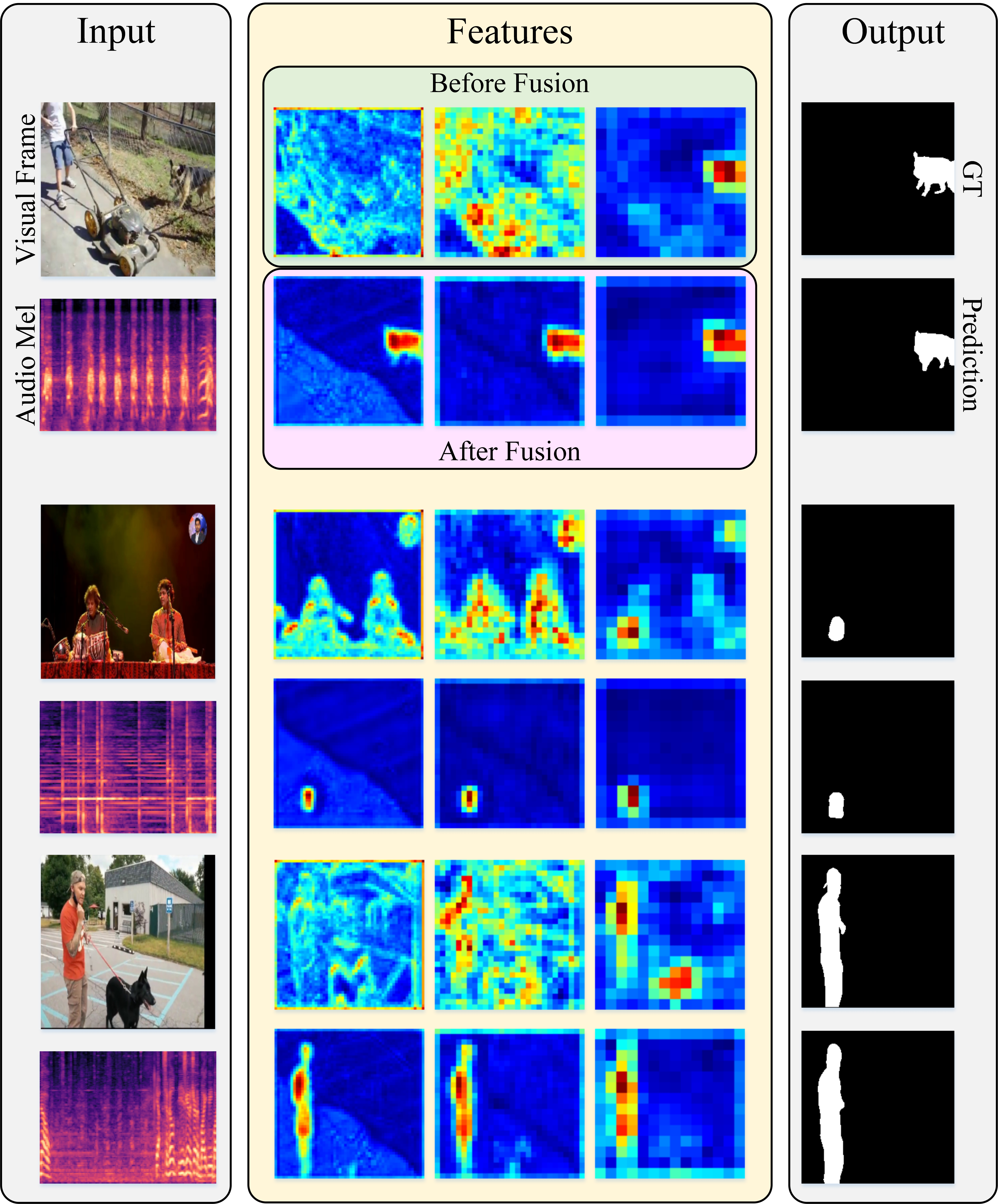}
	\caption{
		Visualization of the intermediate features before and after the fusion of audio and visual information. 
	}
	\label{vis_feat1}
\end{figure}

Fig.~\ref{vis_feat1} illustrates an additional visualization of the intermediate features before and after the fusion. The middle column presents the visual features. The top row displays features derived from visual frames alone (vision-only), while the bottom row shows the fused features after integrating audio signals. The visual features without fusion tend to highlight all objects within the scene indiscriminately. In contrast, the fused features are more focused on regions corresponding to objects associated with the audio signal (e.g., the dog in the first example and the drum in the second). This observation indicates that audio signals act as a guiding mechanism, directing the visual features to concentrate on relevant regions.

\end{document}